\documentclass[runningheads]{llncs}

\usepackage{eccv}

\usepackage{eccvabbrv}
\usepackage{multirow}
\usepackage{colortbl}
\usepackage{wrapfig,lipsum,booktabs}
\usepackage{xcolor}
\usepackage{breqn}
\usepackage{adjustbox} 

\usepackage{comment}
\usepackage{algorithm}
\usepackage[noend]{algpseudocode}
\usepackage[globalcitecopy]{bibunits}

\newcommand*{\V}{\mathbf}
\newcommand{\mr}{MASt3R\xspace}

\definecolor{myemerald}{rgb}{0.753, 0.898, 0.804}
\definecolor{mylightgreen}{rgb}{0.894, 0.933, 0.745}
\definecolor{myyellow}{rgb}{0.996, 0.972, 0.780}

\colorlet{colorFst}{Green!20}       %
\colorlet{colorSnd}{SpringGreen!45} %
\colorlet{colorTrd}{Yellow!30}      %
\colorlet{colorSep}{blue!5}         %
\newcommand{\spacebetweentablerow}{.3}

\newcommand{\argmin}{\mathop{\mathrm{argmin}}} 

\usepackage[accsupp]{axessibility}  

\usepackage{hyperref}

\usepackage{orcidlink}

\begin{document}

\title{SLAM-Former: \\Putting SLAM into One Transformer} 

\titlerunning{SLAM-Former}

\author{Yijun Yuan\inst{1}\orcidlink{0000-0002-6147-4846} \and
Zhuoguang Chen\inst{1}\orcidlink{0009-0003-3037-4417} \and
Kenan Li\inst{1}\orcidlink{0009-0003-4402-6077} \and
Weibang Wang\inst{1}\orcidlink{0009-0000-2462-1475} \and
Minghui Qin\inst{2}\orcidlink{0009-0003-6243-6395} \and
Zhijian Fang\inst{1}\orcidlink{0009-0007-3778-1057} \and
Weicheng Zheng\inst{2}\orcidlink{0009-0004-1570-1544} \and
Hang Zhao\inst{1,2}\orcidlink{0000-0003-1928-7841}
}

\authorrunning{Y.~Yuan et al.}

\institute{IIIS, Tsinghua University 
\and
Shanghai Qi Zhi Institute\\ 
\vspace{2mm}
\color{blue}{\tt\small{\url{https://tsinghua-mars-lab.github.io/SLAM-Former}}}
\\
\color{black}{\tt\small{\{yuanyj, hangzhao\}@mail.tsinghua.edu.cn}}
	\vspace{-2mm}}


\maketitle
\begin{center}
    \includegraphics[width=1.0\linewidth]{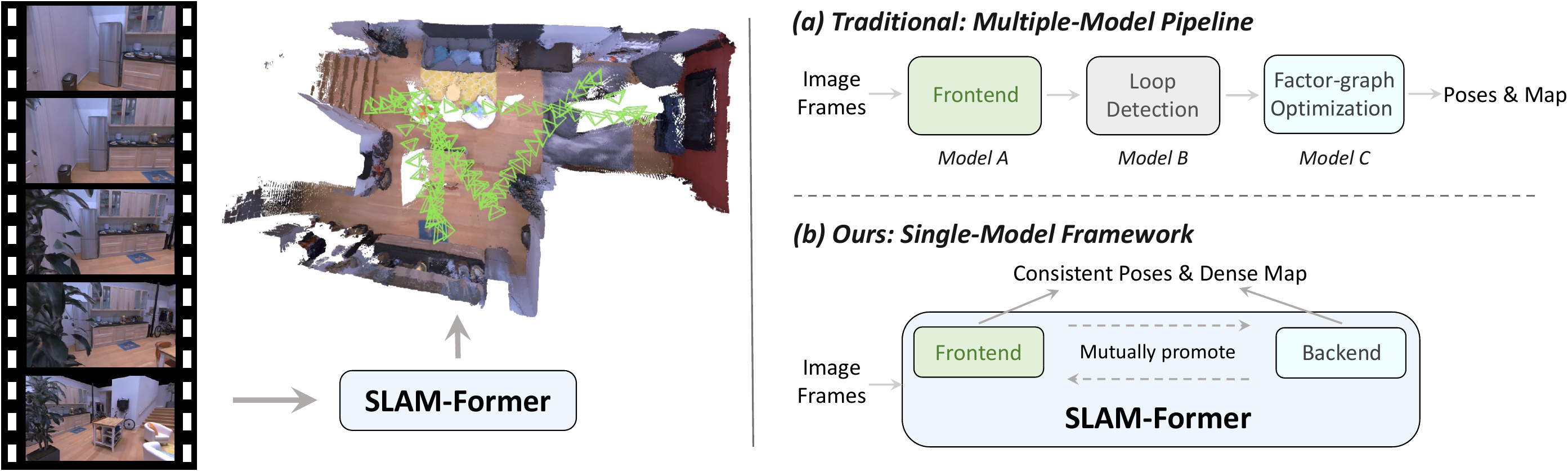}
    \captionof{figure}{SLAM-Former is a unified Transformer for SLAM. Traditional SLAM employs a multi-model pipeline for frontend and backend tasks. In contrast, SLAM-Former integrates the full SLAM functionality within one transformer, achieving consistent poses and dense maps.}
    \label{fig:teaser}
\end{center}







\begin{abstract}
We present SLAM-Former, a neural approach that integrates full SLAM capabilities into a single transformer. 
Similar to traditional SLAM systems, SLAM-Former comprises both a frontend and a backend that operate in tandem. 
The frontend processes sequential monocular images in real-time for incremental mapping and tracking, while the backend performs global refinement to ensure a geometrically consistent result. 
This alternating execution allows the frontend and backend to mutually promote one another, enhancing overall system performance.
Comprehensive experimental results demonstrate that SLAM-Former achieves superior or highly competitive performance compared to state-of-the-art dense SLAM methods.
\end{abstract}  

\section{Introduction}
\label{sec:intro}
In the field of robotic perception, Simultaneous Localization and
Mapping (SLAM) plays a great significance. 
It allows robots to construct a map of an unknown environment while simultaneously
tracking their own location. 
This capability is essential for robots to navigate autonomously and carry out
tasks in a variety of environments. 
Early SLAM algorithms primarily focused on localization and mapping using sparse or semi-dense
representations, such as ORB-SLAM~\cite{Mur_Artal_2015} and LSD-SLAM~\cite{engel2014lsd}. These methods are
efficient and robust, but they may not offer detailed
information about the surroundings.
In contrast, dense mapping techniques aim to create a more detailed and continuous representation of the environment, mainly relying on LiDAR and RGB-D~\cite{elseberg2013one,6162880}.

With the rapid advancement of optical flow and multiview depth estimation techniques, recent research has achieved high-quality dense monocular SLAM
using only images as input~\cite{NEURIPS2021_89fcd07f,yuan2025scenefactoryworkflowcentricunifiedframework,murai2025mast3rslamrealtimedenseslam,maggio2025vggtslamdensergbslam}. These approaches leverage the
capability of neural networks and computer vision algorithms to
estimate depth and motion from a single camera, thereby
creating dense maps without additional sensors.
Especially noteworthy is the trend of utilizing geometric foundation models such as DUSt3R~\cite{wang2024dust3rgeometric3dvision} and VGGT~\cite{wang2025vggtvisualgeometrygrounded}.
These models reveal the high potential of data-driven 3D structure prediction.
Their streaming variants, StreamVGGT~\cite{zhuo2025streaming4dvisualgeometry} and Stream3R~\cite{lan2025stream3rscalablesequential3d}, by carefully leveraging the attention key-value cache (KV cache), enable the model to process incremental visual inputs.

We observe that SLAM methods using geometric foundation models as their reconstruction module, such as MASt3R-SLAM~\cite{murai2025mast3rslamrealtimedenseslam} and VGGT-SLAM~\cite{maggio2025vggtslamdensergbslam} suffer from global inconsistency, since they rely on the alignment of local submaps.
On the other hand, streaming methods such as StreamVGGT and Stream3R process incremental inputs without remapping the past, which can cause a significant mismatch between past and newly incoming data.

In this work, we introduce SLAM-Former, the first fully neural SLAM system with a complete SLAM pipeline, built upon a single unified transformer architecture.
SLAM-Former consists of a frontend and a backend that operate cooperatively, as illustrated in~\cref{fig:teaser} (b).
The frontend operates in real-time on the sequential RGB images for keyframe selection and incremental map and pose updates. 
With the incremental output from the frontend, our backend periodically refines the map and poses globally.
The frontend and backend promote each other. 
The frontend provides initial results and sequential order, assisting the backend in refinement. In return, the backend's KV cache is updated to the frontend periodically 
for further incremental operation.
To empower a single transformer with all SLAM capabilities, we propose three training modes for it.

Compared to traditional SLAM pipelines that require an additional loop detection module to close the pose graph, SLAM-Former's backend accomplishes this with full attention and is equivalent to processing loop detection. 

SLAM-Former achieves significantly better reconstruction and state-of-the-art tracking performance on widely used Dense Mono SLAM benchmarks, compared to both calibrated and uncalibrated state-of-the-art methods.

The main contributions of this work are:
\begin{itemize}
    \item \textbf{Novel SLAM Architecture}. SLAM-Former is the first fully neural SLAM system featuring a complete SLAM pipeline. It introduces a novel Frontend-Backend collaboration mechanism with map tokens and KV caches.
    \item \textbf{Novel Function Basis}. We introduce the first unified transformer formulation for SLAM, 
    jointly training three functional modes within a single transformer to enable complete SLAM functionality.
    \item \textbf{Performance}. SLAM-Former achieves outstanding reconstruction and tracking on widely used benchmarks, demonstrating the high potential of transfor\-mer-based SLAM.
\end{itemize}
 
\section{Related Works}
\subsection{Dense RGB SLAM}
In recent years, research on dense SLAM using monocular cameras has achieved significant progress~\cite{NEURIPS2021_89fcd07f, zhu2023nicerslamneuralimplicitscene,yuan2025scenefactoryworkflowcentricunifiedframework, murai2025mast3rslamrealtimedenseslam,maggio2025vggtslamdensergbslam}, thanks to the application of deep learning techniques. 
Due to the absence of depth sensors, dense RGB SLAM optimizes the entire sequence of geometry and camera as a whole.

Early works focus on reducing the computational cost of depth estimation. For instance,
CodeSLAM~\cite{bloesch2019codeslamlearningcompact} and DeepFactors~\cite{Czarnowski_2020} optimize the depth latent as an alternative.
Borrowing the strength of MVSNet~\cite{yao2018mvsnetdepthinferenceunstructured}, Tandem relies on an external model, but it breaks the co-optimization structure~\cite{koestler2022tandem}.
Conversely, DROID-SLAM~\cite{NEURIPS2021_89fcd07f} and SceneFactory~\cite{yuan2025scenefactoryworkflowcentricunifiedframework} incorporate deep optical flow model into the pipeline and co-optimize both with a dense bundle adjustment.
On the other hand, NeRF~\cite{mildenhall2020nerfrepresentingscenesneural} and Gaussian Splatting~\cite{kerbl20233dgaussiansplattingrealtime} based methods have emerged as a trend to reshape Dense SLAM. 
NeRF-SLAM methods~\cite{rosinol2022nerfslamrealtimedensemonocular,zhu2023nicerslamneuralimplicitscene} and GS-SLAM methods~\cite{yan2024gsslamdensevisualslam} optimize the scene as a whole for a highly realistic novel view synthesis objective.
Nonetheless, those rendering-based SLAMs are time-consuming, do not provide true 3D geometry, and highly sensitive to blur and noise, which restricts their use in real life.

With the emergence of recent foundational geometry techniques, such as DUSt3R~\cite{wang2024dust3rgeometric3dvision} and VGGT~\cite{wang2025vggtvisualgeometrygrounded}, researchers have found new inspiration. MASt3R-SLAM~\cite{murai2025mast3rslamrealtimedenseslam} utilizes the pairwise model MASt3R~\cite{leroy2024groundingimagematching3d} for calibration-free matching and geometry construction, demonstrating state-of-the-art performance under a traditional SLAM pipeline.
On the other hand, VGGT-SLAM~\cite{maggio2025vggtslamdensergbslam} feeds submaps into VGGT and connects them using a novel SL(4) manifold, modeling the geometry distortion from geometric foundational model for the first time.

Existing methods typically rely on pair- or submap-wise geometry optimization, which often leads to structural inconsistencies across frames. MASt3R-SLAM improves cross-frame consistency through global optimization and local TSDF fusion~\cite{6162880}, while VGGT-SLAM introduces sparse cross-submap connections. However, neither method explicitly models global geometric consistency across all frames.

This limitation motivates our proposed SLAM framework, which enforces such consistency during SLAM processing.



\subsection{Feed-forward 3D Reconstruction}
 Recently, DUSt3R~\cite{wang2024dust3rgeometric3dvision} has led a trend to regress the 3D structure with scalable training data directly. However, with processing image pairs, DUSt3R requires a global optimization for larger scenes, which lowers the inference efficiency. 
Several works have been proposed to address this limitation. Fast3R~\cite{yang2025fast3r3dreconstruction1000}, VGGT~\cite{wang2025vggtvisualgeometrygrounded}, and Pi3~\cite{wang2025pi3scalablepermutationequivariantvisual} process multi-view images in a single forward pass, avoiding time-consuming post-processing global optimization. All three are trans\-former-based models for multi-view pointmap estimation. Fast3R highlights the ability to efficiently handle thousands of images, while VGGT demonstrates that a simple architecture with 3D multi-task learning and scalable training data can achieve state-of-the-art results. Pi3 further introduces a permutation-equivariant design that removes the dependence on a fixed reference view, enhancing robustness to input ordering and scalability.


Besides feed-forward multi-view methods, recent feed-forward streaming approaches tackle online 3D reconstruction. Spann3R~\cite{wang20243dreconstructionspatialmemory} extends Dust3R to streaming via an interactive spatial memory, while CUT3R~\cite{wang2025continuous3dperceptionmodel} introduces persistent state tokens with recurrent transformer updates. LONG3R~\cite{chen2025long3rlongsequencestreaming} adopts a 3D spatio-temporal memory and a coarse-to-fine pipeline for long-sequence streaming. Going further, StreamVGGT~\cite{zhuo2025streaming4dvisualgeometry} and STream3R~\cite{lan2025stream3rscalablesequential3d} leverage language-model-inspired causal attention for streaming reconstruction.


However, existing streaming methods focus solely on incremental updates without revisiting past estimates, leading to drift and limited global consistency. To address this, we propose SLAM-Former, a unified neural SLAM pipeline that integrates both frontend and backend for efficient incremental updates and periodic global refinement.

\section{SLAM-Former}
\label{sec:slam}
This section introduces \emph{SLAM-Former}. We first describe the underlying transformer architecture in~\cref{sec:transformer}, then detail its roles in the SLAM frontend (\cref{sec:frontend}) and backend (\cref{sec:backend}). 
Next, we present a joint training strategy that unifies these tasks within a single model in~\cref{sec:training}, and a detailed SLAM inference pipeline in~\cref{sec:execution}, followed by a token pruning technique in \cref{sec:tokenprune}.

\subsection{Transformer Architecture}
\label{sec:transformer}
SLAM-Former is built upon a single transformer model, where a \textbf{transformer backbone $f$} aggregates both intra-frame and inter-frame information, and task-specific \textbf{heads $h$} decode scene geometry and camera poses.  
We assume image features are pre-encoded, and the input to $f$ contains a set of image patch tokens augmented with register tokens. 
The backbone contains $L$ layers with intra-frame and inter-frame attention to jointly capture local image context and temporal correspondences.  



\begin{figure*}[t!]
	\centering
	\includegraphics[width=\linewidth]{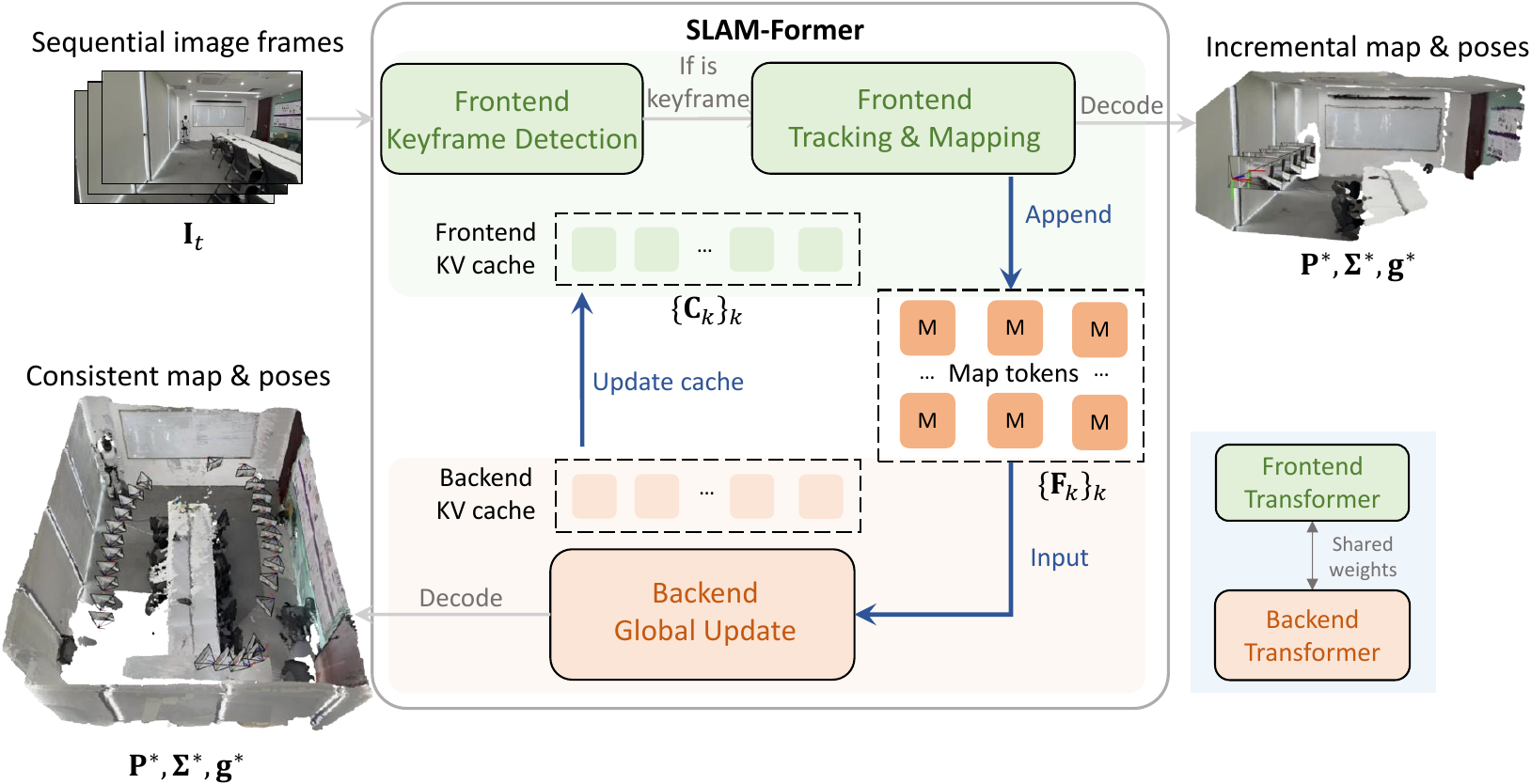} 
	\caption{The working pipeline of SLAM-Former. The frontend detects keyframes and performs incremental pose and map updates, while the backend performs global pose and map updates.
    The shared map token memory and the KV cache update mechanism ensure that the frontend and the backend promote each other, and this process is marked by blue arrows \protect\adjustbox{valign=c}{\includegraphics[height=1\fontcharht\font`A]{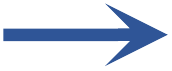}}. }
	\label{fig:slam}
\end{figure*}

SLAM-Former adopts a unified architecture that supports both a \emph{frontend} for incremental frame processing and a \emph{backend} for global map and pose refinement within the same transformer backbone (see~\cref{fig:slam}).

\subsection{Frontend}
\label{sec:frontend}
We illustrate the frontend process in~\cref{fig:slam}.  
When a new frame arrives, the frontend first decides whether it should be a new keyframe.  
If so, the system proceeds with tracking and mapping.  

Formally, given an image sequence $\{\V I_t \in \mathbb{R}^{3\times H\times W}\}_{t\in \{1,2,\cdots\}}$, the \textbf{frontend} $f_{\text{fn}}$ maps each frame to a set of \emph{map tokens}:
\begin{equation}
    \V F_t = f_{\text{fn}}(\V I_t)_{\{\V C_{k}\}_{k\in \V S}}
    \label{eq:f_frontend}
\end{equation}
where $\{\V C_{k}\}_{k\in \V S}$ denotes the \textbf{KV cache} of previous keyframes, storing the key (K) and value (V) tensors from the inter-frame attention. Here, $\V S$ is the set of keyframe indices, and $\V F_t$ denotes the resulting set of map tokens for frame $t$, serving as an implicit neural representation of the scene.  
The newly generated KV cache in this process, $\V C_{t} = \text{Cache}(f(\V F_t))$, is appended to $\{\V C_k\}_{k\in \V S}$ for subsequent use.
Optionally, task-specific heads extract the pointmap $\V P_t$, confidence $\V \Sigma_t$, and camera pose $\V g_t$: $\V P_t, \V \Sigma_t, \V g_t = h(\V F_t)$.

\subsubsection{Keyframe Detection.}
After generating map tokens $\V F_t$, the frontend estimates the camera pose using the pose head $h_{\text{pose}}$:  
$    \V g_t = h_{\text{pose}}(\V F_t)$.
A frame is marked as a new keyframe if its relative pose to the most recent keyframe $k_{\text{prev}}$,
$    \V g_{k_{\text{prev}},t} = \V g_{k_{\text{prev}}}^{-1}\V g_t$,
exceeds a threshold $\tau$.  

In practice, keyframe detection does not rely on the KV cache; instead, we directly apply $f_{\text{fn}}(\V I_{k_{\text{prev}}}, \V I_t)$ to the frame pair \((k_{\text{prev}}, t)\), which improves computational efficiency. 

\subsubsection{Frontend Tracking and Mapping.}
Once a new keyframe is confirmed, $\V F_t$ is recomputed using the full KV cache as defined in~\cref{eq:f_frontend}, and the token map $(\V M, \V S)$ is updated:  
\begin{equation}
    \V M \leftarrow \V M \cup \{\V F_t\}, \quad
    \V S \leftarrow \V S \cup \{t\}.
\end{equation}
The frontend depends only on past frames, making it causal and suitable for online tracking.  
However, this causal design inevitably leads to error accumulation and local inconsistencies.  
To mitigate this, we introduce a \textbf{backend} module for global refinement.

\subsection{Backend}
\label{sec:backend}
The backend is responsible for refining the map tokens to enforce global consistency.  
As illustrated in~\cref{fig:teaser}, traditional SLAM pipelines typically rely on loop closure detection and graph optimization to achieve this objective.  
In contrast, our approach employs a transformer-based \textbf{backend} $f_{\text{bn}}$ that directly refines all map tokens in a single forward pass:
\begin{equation}
    \bar{\V M} = f_{\text{bn}}(\V M).
\end{equation}

The effectiveness of this design stems from the full attention mechanism within $f_{\text{bn}}$, which establishes dense interactions among all map tokens.  
This global receptive field enables the backend to correct accumulated drift and enforce structural coherence throughout the reconstructed scene.  

\subsubsection{Cache Sharing.}
To benefit from the backend refinement, the frontend reuses the backend's shared KV cache $\V C_{\V M}$:  $    \{\V C_{k}\}_{k\in \V S} \leftarrow \V C_{\V M}$.
Consequently, subsequent frames are tracked and mapped with respect to the refined global structure, thereby reducing the risk of error accumulation over long sequences.

\subsection{Training Strategy}

\label{sec:training}
The training strategy is designed to enable a single transformer to support both frontend and backend SLAM functionalities, as illustrated in~\cref{fig:train}.  
We jointly train SLAM-Former in three modes, each corresponding to a distinct input–output configuration, within a single iteration.


Batched training under causal, mixed, and full attention is mathematically equivalent to the corresponding inference configurations: (1) frontend inference with accumulated KV caches, (2) frontend inference with mixed KV caches from backend and frontend, and (3) backend inference with full attention. 

\subsubsection{Training Frontend.} 
\begin{figure*}[t!]
 	\centering
 	\includegraphics[width=.9\linewidth]{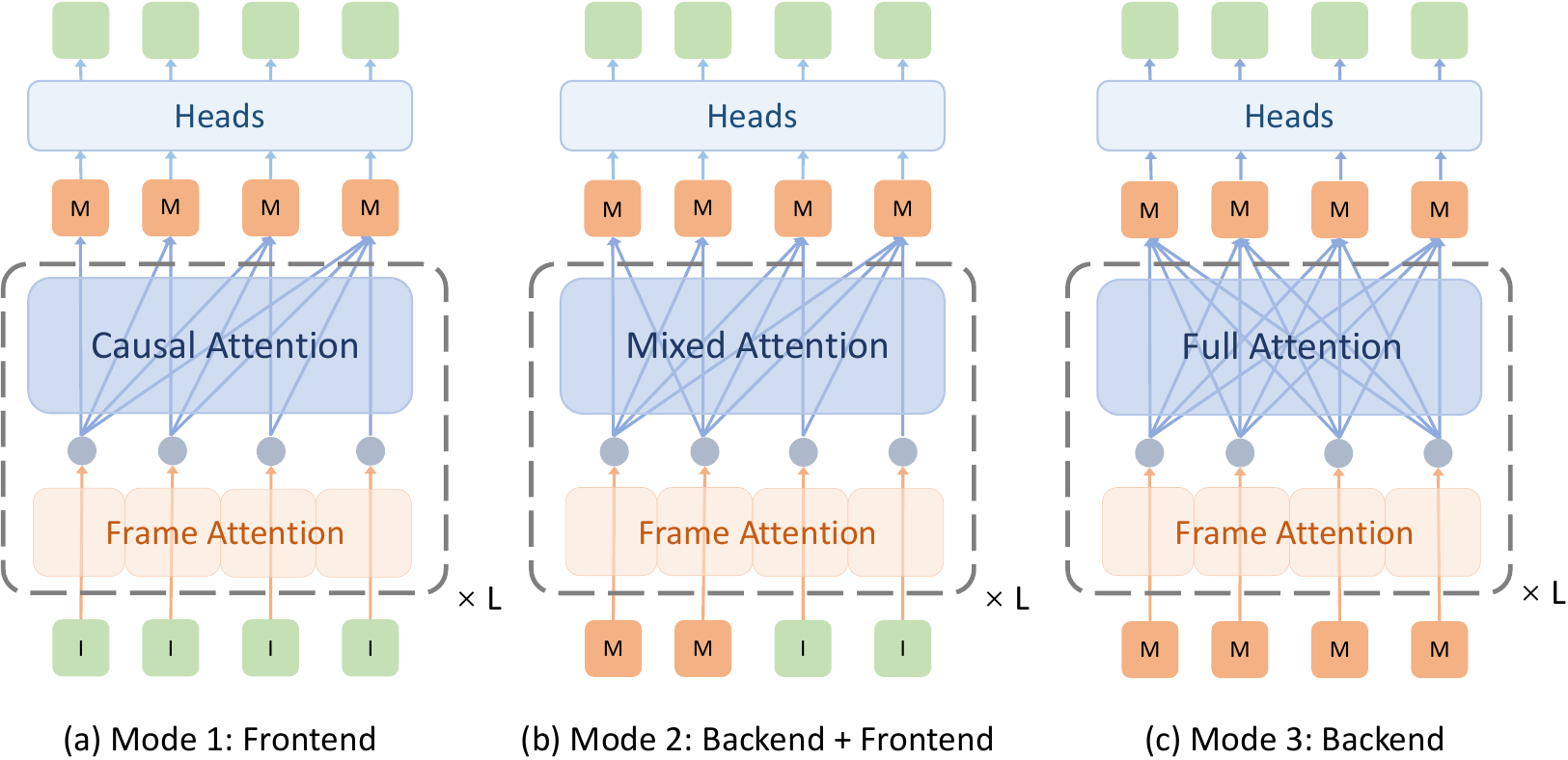}
 	\caption{Three training modes of SLAM-Former. 
    \protect\adjustbox{valign=c}{\includegraphics[height=2\fontcharht\font`A]{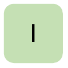}} and 
    \protect\adjustbox{valign=c}{\includegraphics[height=2\fontcharht\font`A]{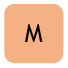}} represent the patch-wise image tokens and map tokens of a frame. 
    \protect\adjustbox{valign=c}{\includegraphics[height=2\fontcharht\font`A]{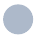}} and 
    \protect\adjustbox{valign=c}{\includegraphics[height=2\fontcharht\font`A]{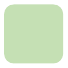}} represent the layer intermediate and final output.
    In each mode, either \protect\adjustbox{valign=c}{\includegraphics[height=2\fontcharht\font`A]{ims/I.pdf}} or 
    \protect\adjustbox{valign=c}{\includegraphics[height=2\fontcharht\font`A]{ims/M.pdf}} tokens, or both, are fed into the transformer backbone $f$, which contains L layers of frame attention and various inter-frame attentions. 
    Finally, pose and pointmap are regressed by the heads $h$.}
 	\label{fig:train}
\end{figure*}

The frontend is trained with a \emph{causal attention} mask (Mode 1 in~\cref{fig:train}~(a)). During inference, it reuses KV caches from previous frames in a mathematically equivariant manner: $ \V F = f(\V I)_{\text{KV}}$.
This enables efficient, end-to-end learning in a single pass.



\subsubsection{Training Frontend with Backend Cooperation.}
To bridge frontend and backend operations (Mode 2), we train $f$ with \emph{mixed attention} to process backend and cache sharing functionality.  
Specifically, the backend refines map tokens with full attention,  
$\bar{\V M} = f_{\text{bn}}(\V M)$,
while the frontend processes new images in the same forward pass as the backend, using causal attention that is equivalent to conditioning on the backend-refined KV cache:
$\V F = f_{\text{fn}}(\V I)_{\V C_{\V M}}$.  

\subsubsection{Training Backend.}
The backend (Mode 3) refines map tokens originating from different runs or KV cache states.  
\emph{Full attention} is applied throughout, allowing the model to correct drift and enforce global consistency:  
    $\bar{\V M} = f_{\text{bn}}(\V M)$.

\subsubsection{Joint Training.}
In all modes, the resulting tokens serve as implicit representations of geometry and camera poses.  
Task-specific heads predict the pointmap $\V P^*$, confidence $\V \Sigma^*$, and camera pose $\V g^*$:  
$\V P^*, \V \Sigma^*, \V g^* = h(\V F)$.
Unlike VGGT, which predicts global geometry, SLAM-Former produces local pointmaps for each frame to avoid the need to define a specific world coordinate.  

The overall loss combines depth, pointmap, and camera supervision: 
$    L = L_{\text{depth}} + L_{\text{pmap}} + \lambda L_{\text{cam}}$.
For depth loss, the predicted depth $\V D^* = \V P^*_z$ is supervised against ground-truth depth $\V D$, weighted by confidence $\Sigma^*$:
\begin{equation}
    L_{\text{depth}} = \sum_t \Big( 
        \|\Sigma^*_t \odot (s^* \V D^*_t - \V D_t)\| 
        + \|\Sigma^*_t \odot (\nabla s^* \V D^*_t - \nabla \V D_t)\| 
        - \alpha \log \Sigma^*_t
    \Big),
\end{equation}
where $\odot$ is element-wise multiplication, $\nabla$ the spatial gradient, and $s^*$ a scale factor estimated following Pi3:
$    s^* = \argmin_s \sum_t \| (s\V P^*_t - \V P_t)/\V D_t \|_1$.
For pointmap loss, similar to depth loss but defined on transformed local pointmaps aligned to the first frame:
$
    \V P^*_{t,1} = {\V g^*_1}^{-1} \V g^*_t \V P^*_t
$, the loss is designed as
\begin{equation}
    L_{\text{pmap}} = \sum_t \Big(
        \|\Sigma^*_t \odot (s^* \V P^*_{t,1} - \V P_t)\|
        + \|\Sigma^*_t \odot (\nabla s^* \V P^*_{t,1} - \nabla \V P_t)\|
        - \alpha \log \Sigma^*_t
    \Big).
\end{equation}

Camera loss employs a scaled Huber loss for relative pose supervision:
\begin{equation}
    L_{\text{cam}} = \sum_{i,j} \|
        s^* \otimes (\V g_i^{*^{-1}} \V g^*_j) -
        (\V g_i^{-1} \V g_j)
    \|_\epsilon,
\end{equation}
where $\otimes$ scales translations and $\|\cdot\|_\epsilon$ denotes the Huber norm.  
All three modes are executed sequentially in a single iteration with shared weights.  
The final training objective is $
    L_{\text{all}} = L_1 + L_2 + \beta L_3$.

\subsection{Execution Pipeline}
\label{sec:execution}
The execution pipeline integrates frontend and backend for online incremental SLAM inference, both powered by the same SLAM-Former transformer.

\subsubsection{Frontend.}
Each incoming frame is first passed to the keyframe detector. Specifically, the current frame and the most recent keyframe are jointly processed by the transformer using full attention to estimate their poses. If the relative spatial distance exceeds a predefined threshold $\tau$, the incoming frame is designated as a new keyframe. 


The first two keyframes are jointly processed for initialization, producing map tokens $\{\mathbf{F}_1, \mathbf{F}_2\}$ and KV caches $\{\mathbf{C}_1, \mathbf{C}_2\}$, which are stored. 
For subsequent keyframes ($k > 2$), the frontend utilizes the KV caches $\{\mathbf{C}_i\}_{i < k}$ to track the $k$-th keyframe, generating the corresponding map tokens $\mathbf{F}_k$ and KV cache $\mathbf{C}_k$.

\subsubsection{Backend.}
Periodically, after every $T$ keyframes (at $k$-th), the accumulated map tokens are refined by the backend, and the resulting KV caches are used to update the frontend KV caches $\{\V C_i\}_{i \leq k}$.  

\subsection{KV Pruning for Runtime Efficiency}
\label{sec:tokenprune}

To alleviate the scalability challenge (latency and VRAM) in long-sequence processing, 
we introduce a KV pruning mechanism to \textbf{speed up backend}. 
Then reuse the pruned KV as cache to \textbf{further accelerate the frontend}.

Inspired by DivPrune~\cite{alvar2025divprune}, we formulate the token retention as a Max-Min Diversity Problem (MMDP). 
For each keyframe, let \(\mathcal{T} = \{\mathbf{t}_1, \dots, \mathbf{t}_{P}\}\) denote the \(P\) non-register patch tokens (and their associated KV pairs). 
We select a subset \(\mathcal{S} \subset \mathcal{T}\) of size \(k = \gamma \cdot P\) that maximizes the minimum pairwise diversity:
\begin{equation}
\mathcal{S}^* = \arg\max_{\substack{\mathcal{S} \subset \mathcal{T}, |\mathcal{S}|=k}} \min_{\substack{\mathbf{t}_i, \mathbf{t}_j \in \mathcal{S}, i \neq j}} d(\mathbf{t}_i, \mathbf{t}_j),
\label{eq:mmdp}
\end{equation}
with pairwise diversity measured by cosine distance $d(\mathbf{t}_i, \mathbf{t}_j) = 1 - {\mathbf{t}_i^\top \mathbf{t}_j}/({\|\mathbf{t}_i\| \|\mathbf{t}_j\|})$. 
We adopt a greedy forward selection algorithm following~\cite{alvar2025divprune} to iteratively select the tokens, achieving near-optimal diversity with \(\mathcal{O}(P^2)\) precomputation cost.

Naively discarding tokens loses the corresponding point cloud in reconstruction. \textbf{Instead, we prune only KV while preserving all queries}. 
For each keyframe,
$
\mathbf{K}' = \mathbf{K}_{[\mathcal{S}^*]}, \quad \mathbf{V}' = \mathbf{V}_{[\mathcal{S}^*]}, \quad \mathbf{Q}' = \mathbf{Q}$,
where \([\mathcal{S}^*]\) indexes the retained tokens.
By sparsifying the keys and values while retaining all queries, the dominant attention computation in the frontend is reduced from $\mathcal{O}(n)$ to $\mathcal{O}(\gamma n)$ under a constant retention ratio, and to \textbf{sublinear ${o}(n)$ under a dynamic retention ratio that decays with sequence length}. Likewise, the backend complexity is reduced from $\mathcal{O}(n^2)$ to $\mathcal{O}(\gamma n^2)$ with a constant retention ratio, and to $\mathcal{O}(n)\cdot{o}(n)$ with the same dynamic retention ratio. This yields $2\times$, $4\times$, $8\times$, or greater speedups (see~\cref{sec:exp:tokenprune}) while maintaining comparable performance.

\section{Experiments}

We evaluate SLAM-Former on multiple tasks, including camera tracking (\cref{sec:exp:track}) and dense 3D reconstruction (\cref{sec:exp:recon}). Subsequently, we analyze the impact of our frontend-backend design (\cref{sec:exp:ablation}), and speedup with KV Pruning (\cref{sec:exp:tokenprune}).

\subsection{Experimental Setup}
\subsubsection{Implementation Details.}
SLAM-Former has 36 transformer layers of both frame- and global-attention in total.
We initialize SLAM-Former with Pi3 pre-trained weights and train $10$ epochs with a batch size of $32$ (excluding the frozen image encoder and camera head).
During training, each batch only contains $12$ frames, while the test is applied on much longer sequences.
For training, we utilize the AdamW optimizer with a learning rate of $1e-5$, and a cosine learning rate scheduler.
In the loss function, the hyperparameters are set as $\lambda=100$ and $\beta=10$.
We set $\tau=0.1$ for translation, backend run every $T=10$ keyframes. 
Regarding datasets, SLAM-Former is trained on ARKitScenes~\cite{baruch2022arkitscenesdiverserealworlddataset}, ScanNet~\cite{dai2017scannet}, ScanNet++~\cite{yeshwanthliu2023scannetpp}, HyperSim~\cite{roberts:2021}, BlendedMVS~\cite{yao2020blendedmvslargescaledatasetgeneralized}, MegaDepth~\cite{MegaDepthLi18}, and MVS-Synth~\cite{DeepMVS}.
During each iteration, all three modes of a single SLAM-Former are trained.
The entire training process takes $11$ hours on $32$ A100 GPUs. Evaluation is conducted on a single RTX 4090 GPU during testing.

\subsubsection{Baselines.}
The baselines utilized in our experiments are categorized as Calibrated and Uncalibrated:
\begin{itemize}
\item Calibrated: ORB-SLAM3~\cite{Campos_2021}, DeepV2D~\cite{teed2020deepv2dvideodepthdifferentiable}, DeepFactors~\cite{Czarnowski_2020}, DPV-SLAM~\cite{lipson2024deeppatchvisualslam}, DPV-SLAM++~\cite{lipson2024deeppatchvisualslam}, GO-SLAM~\cite{zhang2023goslamglobaloptimizationconsistent}, DROID-SLAM~\cite{NEURIPS2021_89fcd07f}, MASt3R-SLAM~\cite{murai2025mast3rslamrealtimedenseslam} and NICER-SLAM~\cite{zhu2023nicerslamneuralimplicitscene}.  
\item Uncalibrated: DROID-SLAM, MASt3R-SLAM, VGGT-SLAM, EC3R-SLAM~\cite{ec3r-slam}, ViSTA-SLAM~\cite{vista-slam} and SLAM3R~\cite{liu2025slam3rrealtimedensescene}, as well as our method. 
We also test related methods: CUT3R~\cite{wang2025continuous3dperceptionmodel} and StreamVGGT~\cite{zhuo2025streaming4dvisualgeometry} using our keyframes.
\end{itemize}

\noindent Unless otherwise specified, evaluation is performed using all KVs.

\begin{table*}[t!]
	\centering
\caption{Root mean square error~(RMSE) of absolute trajectory error~(ATE) on TUM RGB-D~\cite{sturm12iros} (unit: m).
		The * symbol indicates that the baseline is evaluated in the uncalibrated mode from the VGGT-SLAM paper~\cite{maggio2025vggtslamdensergbslam}, the $+$ symbol indicates that the baseline is tested on our machine.
		}\label{tab:tum_ate}
        \vspace{-3mm}
    \resizebox{1\textwidth}{!}{
	\scriptsize
      \renewcommand{\arraystretch}
      {\spacebetweentablerow}
	\begin{tabular}{l|lcccccccccc} 
		\toprule
		\multirow{2}{*}{} & \multirow{2}{*}{Method} & \multicolumn{9}{c}{Sequence} &  \multirow{2}{*}{Avg}  \\ \cmidrule(lr){3-11}
		& &\texttt{360} &\texttt{desk} &\texttt{desk2} &\texttt{floor} &\texttt{plant} &\texttt{room } &\texttt{rpy} &\texttt{teddy} &\texttt{xyz} & \\
		\midrule
		
		\multirow{8}{*}{\rotatebox[origin=c]{90}{Calib.\ \ \ \ }}
		&  ORB-SLAM3~\cite{Mur_Artal_2015} & $  \times$ &  {0.017} &  0.210 &$  \times$ &  0.034 &$   \times$ &  $\times$ &   $\times$ &  \textbf{0.009} &   N/A \\
		&  DeepV2D~\cite{teed2020deepv2dvideodepthdifferentiable} &  0.243 &  0.166 &  0.379 &  1.653 &  0.203 &  0.246 &  0.105 &  0.316 &  0.064 &  0.375 \\
		&  DeepFactors~\cite{Czarnowski_2020} &  0.159 &  0.170 &  0.253 &  0.169 &  0.305 &  0.364 &  0.043 &  0.601 &  0.035 &  0.233 \\
		&  DPV-SLAM~\cite{lipson2024deeppatchvisualslam} &  0.112 &  0.018 &  0.029 &  0.057 &  0.021 &  0.330 &  0.030 &  0.084 &  {0.010} &  0.076 \\
		&  DPV-SLAM++~\cite{lipson2024deeppatchvisualslam} &  0.132 &  0.018 &  0.029 &  0.050 &  0.022 &  0.096 &  0.032 &  0.098 &  {0.010} &  0.054 \\
		&  GO-SLAM~\cite{zhang2023goslamglobaloptimizationconsistent} &  0.089 &  \textbf{0.016} &  {0.028} &  {0.025} &  0.026 &  {0.052} &  \textbf{0.019} &  0.048 &  {0.010} &  {0.035} \\
		&  DROID-SLAM~\cite{NEURIPS2021_89fcd07f} &  0.111 &  0.018 &  0.042 &  \textbf{0.021} &  \textbf{0.016} &  \textbf{0.049} &  {0.026} &  0.048 &  0.012 &  0.038 \\
		&  \mr-SLAM~\cite{murai2025mast3rslamrealtimedenseslam} &  \textbf{0.049} &  \textbf{0.016} &  \textbf{0.024} &  {0.025} &  {0.020} &  0.061 &  0.027 &  \textbf{0.041} &  \textbf{0.009} &  \textbf{0.030} \\
		\midrule
		\multirow{7}{*}{\rotatebox[origin=c]{90}{Uncalib.}} 
		&CUT3R$^+$~\cite{wang2025continuous3dperceptionmodel}  & 0.102& 0.054& 0.118& 0.211 &0.083 &0.264 &0.044 &0.120 &0.020 &   0.113  \\
        &StreamVGGT$^+$~\cite{zhuo2025streaming4dvisualgeometry} & 0.088&  0.063 & 0.105&  0.604&  0.070 & 0.633&  0.025 & 0.081 & 0.015 & 0.187\\
        &DROID-SLAM*~\cite{NEURIPS2021_89fcd07f} &0.202 &  0.032 &0.091 &  {0.064} &0.045 &0.918 &0.056 &0.045 &  0.012 &0.158 \\
		&\mr-SLAM*~\cite{murai2025mast3rslamrealtimedenseslam} & {0.070} & 0.035 &  0.055 &  0.056 &0.035 & 0.118 &0.041 &0.114 & 0.020 &  0.060 \\
		&VGGT-SLAM~\cite{maggio2025vggtslamdensergbslam}&  0.071 &  0.025 &  0.040 & 0.141 &  0.023 &  0.102 &  0.030 &  0.034 &  0.014 &  0.053 \\
        & EC3R-SLAM ~\cite{ec3r-slam}& 0.101& 0.038& 0.050& \textbf{0.055}& 0.097& 0.101& 0.045& 0.125 &0.018& 0.070\\
        & ViSTA-SLAM ~\cite{vista-slam} & 0.104 & 0.030 & 0.030 & 0.070 & 0.052 & \textbf{0.067} & 0.023 & 0.080 & 0.015 & 0.052\\
            &\textbf{Ours} & \textbf{0.067} &\textbf{0.018}& \textbf{0.026 }& 0.079 &\textbf{0.021}& {0.082}& \textbf{0.017}& \textbf{0.030} &\textbf{0.011}& \textbf{0.039}\\
            
		\bottomrule
	\end{tabular}
    }

\end{table*}

\subsection{3D Tracking Evaluation}
\label{sec:exp:track}
We first evaluate the tracking performance of SLAM-Former on the TUM RGB-D, 7-Scenes, and Replica. We compute the Root Mean Square Error (RMSE) of Absolute Trajectory Error (ATE) for various methods under both calibrated and uncalibrated settings. 
\subsubsection{TUM RGB-D Tracking.}
In the TUM test, evaluations are conducted on a widely used subset of scenes. 
The results are summarized in~\cref{tab:tum_ate}. As shown, our model consistently outperforms most baselines in the uncalibrated setting.
The superior performance in these more complex sequences, such as the room and floor that involve significant camera rotation and potential for loop closure, suggests that our backend's global refinement is particularly effective at mitigating accumulated drift.
More importantly, it significantly reduced the error relative to calibrated baselines, achieving a highly competitive level.

\begin{table*}[t!]
	\centering
        \caption{RMSE of ATE on 7-Scenes~\cite{glocker2013real-time} (unit: m).
		 The~*~symbol indicates that the baseline is evaluated in the uncalibrated mode from the VGGT-SLAM paper~\cite{maggio2025vggtslamdensergbslam},
         the $+$ symbol indicates that the baseline is tested on our machine.
	}\label{tab:7scenes_ate}
            \vspace{-3mm}

    \resizebox{0.8\textwidth}{!}{
	\scriptsize
      \renewcommand{\arraystretch}{1} 
      \small

	\begin{tabular}{c|lcccccccc} 
		\toprule
		\multirow{2}{*}{} & \multirow{2}{*}{Method} & \multicolumn{7}{c}{Sequence} &  \multirow{2}{*}{Avg}  \\ \cmidrule(lr){3-9}
		& &\texttt{chess} &\texttt{fire} &\texttt{heads} &\texttt{office} &\texttt{pumpkin} &\texttt{kitchen} &\texttt{stairs} & \\
		\midrule
		
		\multirow{3}{*}{\rotatebox[origin=c]{90}{Calib.}} 
		&  NICER-SLAM~\cite{zhu2023nicerslamneuralimplicitscene} &  \textbf{0.033} &  0.069 &  0.042 &  0.108 &  0.200 &   \textbf{0.039} &  0.108 &  0.086 \\
		&  DROID-SLAM~\cite{NEURIPS2021_89fcd07f} &  {0.036} &  {0.027} &  {0.025} &  \textbf{0.066} &  0.127 &   {0.040} &   {0.026} &   {0.049} \\
		&  \mr-SLAM~\cite{murai2025mast3rslamrealtimedenseslam} &  0.053 &  \textbf{0.025} &   \textbf{0.015} &  {0.097} &   \textbf{0.088} &  0.041 &   \textbf{0.011} &   \textbf{0.047} \\ \midrule
        
		\multirow{7}{*}{\rotatebox[origin=c]{90}{\ \ \ Uncalib.}} 

		& CUT3R$^+$~\cite{wang2025continuous3dperceptionmodel} & 0.046& 0.043 &0.055 &0.120& 0.096 &0.061& 0.086& 0.073	\\
            & StreamVGGT$^+$~\cite{zhuo2025streaming4dvisualgeometry} &0.048& 0.036 &0.030&  0.117&  0.094&  0.063&  0.179&  0.081\\
		&DROID-SLAM*~\cite{NEURIPS2021_89fcd07f} & 0.047 & 0.038 & 0.034 & 0.136 & 0.166 & 0.080 &   0.044 & 0.078  \\
		&\mr-SLAM*~\cite{murai2025mast3rslamrealtimedenseslam} & 0.063 & 0.046 & 0.029 &  0.103 &  {0.114} &0.074 &  0.032 &  0.066 \\
            &VGGT-SLAM~\cite{wang2025vggtvisualgeometrygrounded} &  \textbf{0.036} &  \textbf{0.028} & \textbf{ 0.018} &  0.103 &  0.133 & 0.058 & 0.093 &  0.067  \\
            &EC3R-SLAM~\cite{ec3r-slam} & 0.050 &0.042& 0.059& 0.113& 0.143& 0.065& 0.050& 0.075 \\
            &ViSTA-SLAM~\cite{vista-slam} &0.073& 0.035 &0.028 &0.055& 0.129& 0.035 &0.029 &0.055 \\
            & Ours & 0.039&0.033&\textbf{0.018}&\textbf{0.070}&\textbf{0.065}&\textbf{0.035}&\textbf{0.033}&\textbf{0.042}\\
		\bottomrule
	\end{tabular}
    }
        \vspace{0.7em}

\end{table*}

\subsubsection{7-Scenes Tracking.}
We evaluate the tracking performance on 7-Scenes in ~\cref{tab:7scenes_ate}, where our method outperforms most baselines under uncalibrated and calibrated settings.
In more complex scenes, such as the office, pumpkin and kitchen, our model achieves a notably higher performance gap compared to other methods. 
On average, our method outperforms all baselines.

\subsubsection{Replica Tracking.}
The previous tracking experiments were conducted with real captured data, while the replica dataset is synthetic.
Our approach shows substantial improvements under the uncalibrated setting, achieving an approximate $50$\% reduction in ATE compared to SLAM3R and outperforming all baselines, as shown in~\cref{tab:replica_ate}.
However, our method performs on par with NICER-SLAM but still lags behind the traditional SLAM method, DROID-SLAM. This is because synthetic data lacks noise and blur, making the matchings sufficiently accurate for pose solving in bundle adjustment. In contrast, in the previous real-life data tests, DROID-SLAM performed on a similar level to our method.

\begin{table}[t!]
	\centering
        \caption{RMSE of ATE on Replica~\cite{straub2019replicadatasetdigitalreplica} (unit: m).
                 The $+$ symbol indicates that the baseline is tested on our machine.
	}\label{tab:replica_ate}
        \vspace{-3mm}

    \resizebox{0.8\textwidth}{!}{
          \renewcommand{\arraystretch}{1.0} 
	\small
        \begin{tabular}{c|lccccccccc} 
		\toprule
		\multirow{2}{*}{} & \multirow{2}{*}{Method} & \multicolumn{8}{c}{Sequence} &  \multirow{2}{*}{Avg}  \\ \cmidrule(lr){3-10}
    & &\texttt{Rm0} &\texttt{Rm1} &\texttt{Rm2} &\texttt{Of0} &\texttt{Of1} &\texttt{Of2} & \texttt{Of3} &\texttt{Of4}&\\
		\midrule
		\multirow{2}{*}{\rotatebox[origin=c]{90}{Calib.}} 
		&  NICER-SLAM~\cite{zhu2023nicerslamneuralimplicitscene} &  0.013 & 0.016 & 0.011 & 0.021 & 0.032 & 0.021 & 0.014 & 0.020 & 0.019 \\[0.1ex]
		&  DROID-SLAM~\cite{NEURIPS2021_89fcd07f} & \bf  0.003 & \bf 0.001 & \bf 0.003 & \bf 0.003 & \bf 0.004 &  \bf0.003 & \bf 0.005 &\bf  0.004 & \bf 0.003  \\[0.1ex] 
        \midrule
		\multirow{5}{*}{\rotatebox[origin=c]{90}{Uncalib.}} 
            & SLAM3R~\cite{liu2025slam3rrealtimedensescene} &0.046& 0.059& 0.057& 0.112& 0.063& 0.062& 0.050& 0.081& 0.066\\
            & CUT3R $^+$ ~\cite{wang2025continuous3dperceptionmodel}& 0.145& 0.243& 0.127& 0.159& 0.230 &0.162& 0.088& 0.204& 0.170 \\
            & StreamVGGT$^+$~\cite{zhuo2025streaming4dvisualgeometry}  & 0.113&  0.163&  0.077&  0.076&  0.070&  0.180 & 0.153&  0.168&  0.125\\
            & VGGT-SLAM $^+$~\cite{maggio2025vggtslamdensergbslam} &0.030& 0.167& 0.086 & 0.042 & 0.064 & 0.095 & 0.039 & 0.043 &0.071 \\
            & DROID-SLAM~\cite{wang2024dust3rgeometric3dvision} & 0.313 & 0.111 & 0.125  &0.029  &0.421 & 0.045  &0.253  &0.249 & 0.193 \\
            & \mr-SLAM~\cite{murai2025mast3rslamrealtimedenseslam} & 0.023 & 0.035 & 0.103 & 0.038 & 0.033  &0.047 & 0.035 & 0.045 & 0.045 \\
            & EC3R-SLAM~\cite{ec3r-slam} & 0.038 & 0.049 & 0.027 & 0.043  &0.034 & 0.059 & 0.025 & 0.053 & 0.041\\
            & ViSTA-SLAM~\cite{vista-slam} & 0.069& 0.093& 0.136& 0.074& 0.193& 0.118& 0.049& 0.130 & 0.108\\
            & Ours & \bf0.030 &\bf 0.026 &\bf 0.027 &\bf 0.028 &\bf0.029 & \bf0.038&\bf0.028&\bf0.031& \bf 0.030\\
		\bottomrule
	\end{tabular}
    }
\end{table}

\subsection{Reconstruction Evaluation}
\label{sec:exp:recon}
We evaluate the reconstruction performance of our SLAM-Former on the 7-Scenes dataset following the protocol of VGGT-SLAM and on the Replica dataset following the protocol of SLAM3R.

\begin{table}[t!]
	\centering
    	\caption{Reconstruction evaluation on 7-Scenes~\cite{glocker2013real-time} (unit: m). $@n$ indicates a keyframe every $n$ images.}
	\label{tab:7scenes_recon}
        \vspace{-3mm}

	\setlength{\tabcolsep}{2.0pt}
	\scriptsize
\renewcommand{\arraystretch}{\spacebetweentablerow}

	\begin{tabular}{l|lcccc}
		\toprule
		\multirow{2}{*}{} & \multirow{2}{*}{Method} & \multicolumn{4}{c}{7-Scenes} \\ \cmidrule(lr){3-6}
		& & ATE\,$\downarrow$ & Acc.\,$\downarrow$ & Complet.\,$\downarrow$ & Chamfer\,$\downarrow$ \\
		\midrule
		 \parbox[t]{2mm}{\multirow{4}{*}{\rotatebox[origin=c]{90}{Calib.}}}
		&  DROID-SLAM~\cite{NEURIPS2021_89fcd07f} &  0.049 &  0.141 &  0.048 &  0.094  \\
		&  \mr-SLAM~\cite{murai2025mast3rslamrealtimedenseslam} &  {0.047} &  0.089 &  0.085 &  0.087  \\
		&  Spann3R @20~\cite{wang20243dreconstructionspatialmemory}  &  N/A &  0.069 &  0.047 &  0.058 \\ 
		&  Spann3R @2~\cite{wang20243dreconstructionspatialmemory}   &  N/A &  0.124 &  {0.043} &  0.084 \\
		\midrule
		\parbox[t]{2mm}{\multirow{6}{*}{\rotatebox[origin=c]{90}{Uncalib.}}}
        & CUT3R$^+$~\cite{wang2025continuous3dperceptionmodel} & 0.073& 0.032& 0.047& 0.040\\
        &StreamVGGT$^+$~\cite{zhuo2025streaming4dvisualgeometry}&0.081 & 0.058& 0.057& 0.057\\
        & \mr-SLAM*~\cite{murai2025mast3rslamrealtimedenseslam}  &  0.066 &  0.068 &  0.045 & 0.056  \\
		& VGGT-SLAM~\cite{maggio2025vggtslamdensergbslam} &  0.067 &  {0.052} &  0.058 &  {0.055} \\
        & EC3R-SLAM~\cite{ec3r-slam}  & 0.075& 0.025& 0.054& 0.040\\
        & ViSTA-SLAM ~\cite{vista-slam}& 0.056& 0.051& 0.055& 0.045\\
            & \textbf{Ours} & \textbf{ 0.042}& \textbf{0.017}&\textbf{0.037} & \textbf{0.027}\\
		\bottomrule
	\end{tabular}
    \vspace{0.7em}

\end{table}

\begin{table*}[t!]
\caption{
Reconstruction results on the Replica~\cite{straub2019replicadatasetdigitalreplica} dataset.
* denotes the results reported in NICER-SLAM~\cite{zhu2023nicerslamneuralimplicitscene}.
$^-$ shows the results from SLAM3R~\cite{liu2025slam3rrealtimedensescene}.
$^+$ represented the results from our run.
}
\label{tab:replica_rec}
        \vspace{-3mm}

\centering
\resizebox{1\textwidth}{!}{
\renewcommand{\arraystretch}{\spacebetweentablerow}

\small 

\begin{tabular}{l|cccccccc|c}
\toprule

\multirow{2}{*}{Method}& Room 0 & Room 1 & Room 2 & Office 0 & Office 1 & Office 2 & Office 3 & Office 4 & Average  \\ 

& Acc. / Comp. & Acc. / Comp. & Acc. / Comp. & Acc. / Comp. & Acc. / Comp. & Acc. / Comp. & Acc. / Comp. & Acc. / Comp. & Acc. / Comp.  \\ 

\toprule

{DUSt3R}~\cite{wang2024dust3rgeometric3dvision} & 3.47 / {2.50} & {2.53} / {1.86} & {2.95} / {1.76} & 4.92 / 3.51 & {3.09} / {2.21} & 4.01 / 3.10 & {3.27} / {2.25} & 3.66 / 2.61 & {3.49} / {2.48} \\ 

MASt3R~\cite{leroy2024groundingimagematching3d} & 4.01 / 4.10 & 3.61 / 3.25 & 3.13 / 2.15 & {2.57} / {1.63} & 12.85 / 8.13 & {3.13} / {1.99} & 4.67 / 3.15 & 3.69 / {2.47} & 4.71 / 3.36  \\

NICER-SLAM*~\cite{zhu2023nicerslamneuralimplicitscene} & {2.53} / 3.04 & 3.93 / 4.10 & 3.40 / 3.42 & 5.49 / 6.09 & 3.45 / 4.42 & 4.02 / 4.29 & 3.34 / 4.03 & {3.03} / 3.87 & 3.65 / 4.16 \\

DROID-SLAM*~\cite{NEURIPS2021_89fcd07f} & 12.18 / 8.96 & 8.35 / 6.07 & 3.26 / 16.01 & {3.01} / 16.19 & {2.39} / 16.20 & 5.66 / 15.56 & 4.49 / 9.73 & 4.65 / 9.63 & 5.50 / 12.29  \\

DIM-SLAM*~\cite{li2023densergbslamneural} & 14.19 / 6.24 & 9.56 / 6.45 & 8.41 / 12.17 & 10.16 / 5.95 & 7.86 / 8.33 & 16.50 / 8.28 & 13.01 / 6.77 & 13.08 / 8.62 & 11.60 / 7.85 \\

GO-SLAM$^-$~\cite{zhang2023goslamglobaloptimizationconsistent} & - & - & - & - & - & - & - & - & 3.81 / 4.79  \\ 

Spann3R$^-$~\cite{wang20243dreconstructionspatialmemory} & 9.75 / 12.94 & 15.51 / 12.94  & 7.28 / 8.50 & 5.46 / 18.75 & 5.24 / 16.64 & 9.33 / 11.80 & 16.00 / 9.03 & 13.97 / 16.02 & 10.32 / 13.33 \\

{SLAM3R}$^-$~\cite{liu2025slam3rrealtimedensescene} & {3.19} / {2.40} & {3.12} / 2.34 & {2.72} / {2.00} & 4.28 / 2.60 & 3.17 / {2.34} & {3.84} / {2.78} & {3.90} / 3.16 & {4.32} / {3.36} & {3.57} / {2.62}  \\

CUT3R$^+$~\cite{wang2025continuous3dperceptionmodel}&6.09 / 3.09 &
9.89 / 4.55 &
5.77 / 2.66 &
5.23 / 2.46 &
11.60 / 6.94 &
8.00 / 3.16 &
6.12 / 3.09 &
7.46 / 3.05 & 7.52 /3.62\\

StreamVGGT$^+$~\cite{zhuo2025streaming4dvisualgeometry} &
9.01 / 4.37 &
12.22 / 4.66 &
5.61 / 2.61 &
6.64 / 3.45 &
5.14 / 2.55 &
12.84 / 7.23 &
12.09 / 6.59 &
15.48 / 6.35 &
9.88  4.73\\

ViSTA-SLAM~\cite{vista-slam} &7.76 /2.09& 11.05 /5.52& 11.94/ 2.23& 5.95 /2.26& 26.14 /18.33 &9.14/3.16 &6.23/ 2.21& 13.38/ 6.88 &11.45/ 5.34\\

VGGT-SLAM$^+$~\cite{maggio2025vggtslamdensergbslam} &3.23 / 2.66 &
18.92 / 15.41 &
15.93 / 12.37 &
4.01 / 2.49 &
3.01 / 2.32 &
6.93 / 5.31 &
2.97 / 2.53 &
5.19 / 3.83 &
7.52 / 5.86 \\

\mr-SLAM~\cite{murai2025mast3rslamrealtimedenseslam} &2.55 / 2.01 & 2.66  /1.83& 2.14  /1.58 & 2.88 / 1.87 & 3.45 /2.57  &2.80 / 2.06  &3.75 / 2.76  &3.13  /2.21 & 2.92 / 2.25\\
EC3R-SLAM~\cite{ec3r-slam} & 2.07 / 1.63 & 2.90 / 2.03 & 1.93 / 1.57  &4.06  /2.48  &2.17 / 1.69 & 3.41 / 2.43  &2.51 / 2.08  &3.52 / 2.66 & 2.82 / 2.07\\

\textbf{SLAM-Former (Ours)} &\textbf{2.40} / \textbf{1.86} &
\textbf{1.79} / \textbf{1.35} &
\textbf{1.83} /\textbf{ 1.39 }&
\textbf{1.84} / \textbf{1.37} &
\textbf{1.70} / \textbf{1.28} &
\textbf{2.44} /\textbf{ 1.67} &
\textbf{2.36} /\textbf{ 1.89} &
\textbf{2.38 }/ \textbf{1.70} &\textbf{2.09} / \textbf{1.56 }
\\
\bottomrule
\end{tabular}
}
\vspace{0.7em}

\end{table*}
\begin{figure*}[t!]
    \centering
    \includegraphics[width=.95\linewidth]{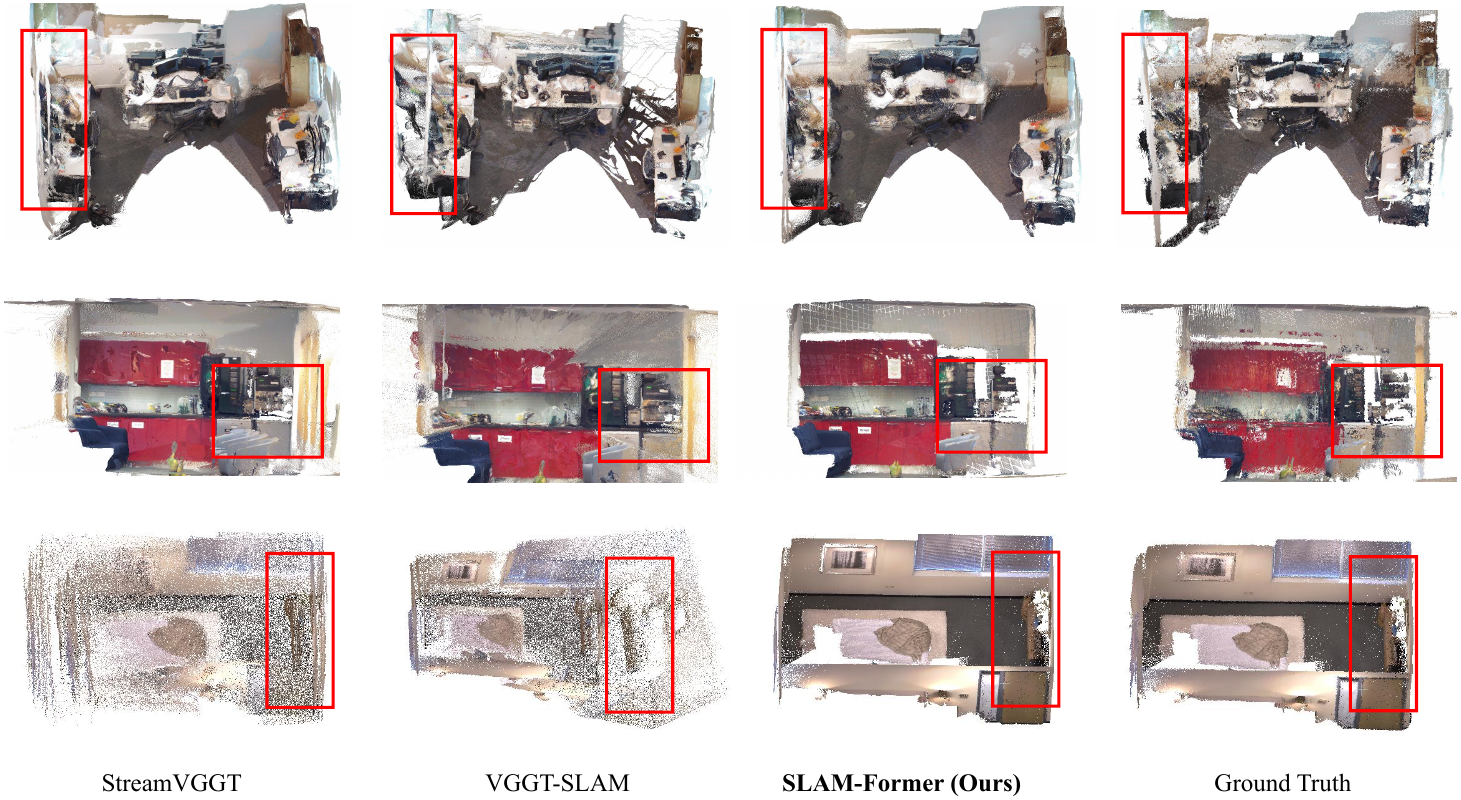} 
    \caption{Qualitative reconstruction comparison. Note the significant structural errors, such as misalignments, from baseline methods (red boxes), which are corrected by SLAM-Former's globally consistent refinement.}
    \label{fig:pdf_scaled}
\end{figure*}

\subsubsection{7-Scenes Reconstruction.}
The reconstruction results on 7-Scenes are shown in~\cref{tab:7scenes_recon}. Our method exhibits a significant gap compared to other dense SLAM state-of-the-art methods. In terms of reconstruction quality, our method achieves the highest accuracy of $0.017$m, while the other methods have an error that is about $47\%$ higher than ours.
In terms of completeness and chamfer distance, our method achieves $0.037$m and $0.027$m, respectively, still outperforming all baselines by around $30\%$.

This consistently superior performance across all major reconstruction metrics is also demonstrated in our reconstruction demo~\cref{fig:pdf_scaled}. 
As shown in the first two rows (office and pumpkin), the baselines exhibit mismatched surfaces between frames within the red-windowed regions. 
In contrast, our SLAM-Former's reconstruction consistently shows coherent and accurate structure.

\subsubsection{Replica Reconstruction.}
The dense reconstruction results on the Replica dataset are listed in~\cref{tab:replica_rec}.
Our method also achieves the best in terms of both accuracy and completion across all baselines.
Specifically, our $2.09/1.56$ Acc./Comp. show large gap to the second best on both metrics.

We also demonstrate the reconstruction in the third row of~\cref{fig:pdf_scaled}. 
Here, StreamVGGT shows multiple layers of surface in the room, as highlighted in the red-windowed region.
More severely, VGGT-SLAM exhibits layers with substantial scale discrepancies.
While SLAM-Former closely matches the ground truth.
The less dense point clouds of the baselines are caused by the mismatch of layers, as the test samples a constant number of points for evaluation.
We also add qualitative comparison with \mr-SLAM in Supplementary Fig. C1.

\subsection{Frontend and backend co-operation}
\label{sec:exp:ablation}
To investigate how the backend design of our SLAM-Former contributes to the overall system performance in a SLAM-like manner, we conducted a series of ablation experiments. 
The results are summarized in~\cref{tab:ablation_tum_ate}. 
All evaluations are conducted on the TUM RGB-D using the RMSE of ATE as the metric.
\begin{table}[htbp]
	\centering
    \caption{Evaluation of module cooperation on TUM RGB-D~\cite{sturm12iros} (unit: m).
		}\label{tab:ablation_tum_ate}
        \vspace{-3mm}
	\scriptsize
        \setlength{\tabcolsep}{1pt}
	\resizebox{.7\textwidth}{!}{
	\renewcommand{\arraystretch}{1}     %
	\begin{tabular}{lcccccccccc} 
		\toprule
		 \multirow{2}{*}{Method} & \multicolumn{9}{c}{Sequence} &  \multirow{2}{*}{Avg}  \\ \cmidrule(lr){2-10}
		 &\texttt{360} &\texttt{desk} &\texttt{desk2} &\texttt{floor} &\texttt{plant} & \texttt{room } &\texttt{rpy} &\texttt{teddy} &\texttt{xyz} & \\
         \midrule
            Frontend-only &0.137& 0.041& 0.045& 0.264& 0.053&  0.547& 0.036& 0.073& 0.013& 0.134\\
            \midrule
            Frontend+Backend & \textbf{0.067} &\textbf{0.018}& \textbf{0.026 }&\textbf{ 0.079} &\textbf{0.021}&  \textbf{0.082}& \textbf{0.017}& \textbf{0.030} &\textbf{0.011}& \textbf{0.039}\\
		\bottomrule
	\end{tabular}
        }
\end{table}

The results demonstrate that the inclusion of backend modules yields significantly improved accuracy over using the frontend alone, confirming the effectiveness of our proposed frontend-backend design. 

In addition to the above quantitative experiments, we also analyze how the frontend and backend mutually promote each other.


\subsubsection*{How does the backend assist the frontend?}
\begin{figure*}[t!]
	\centering
    \includegraphics[width=\linewidth]{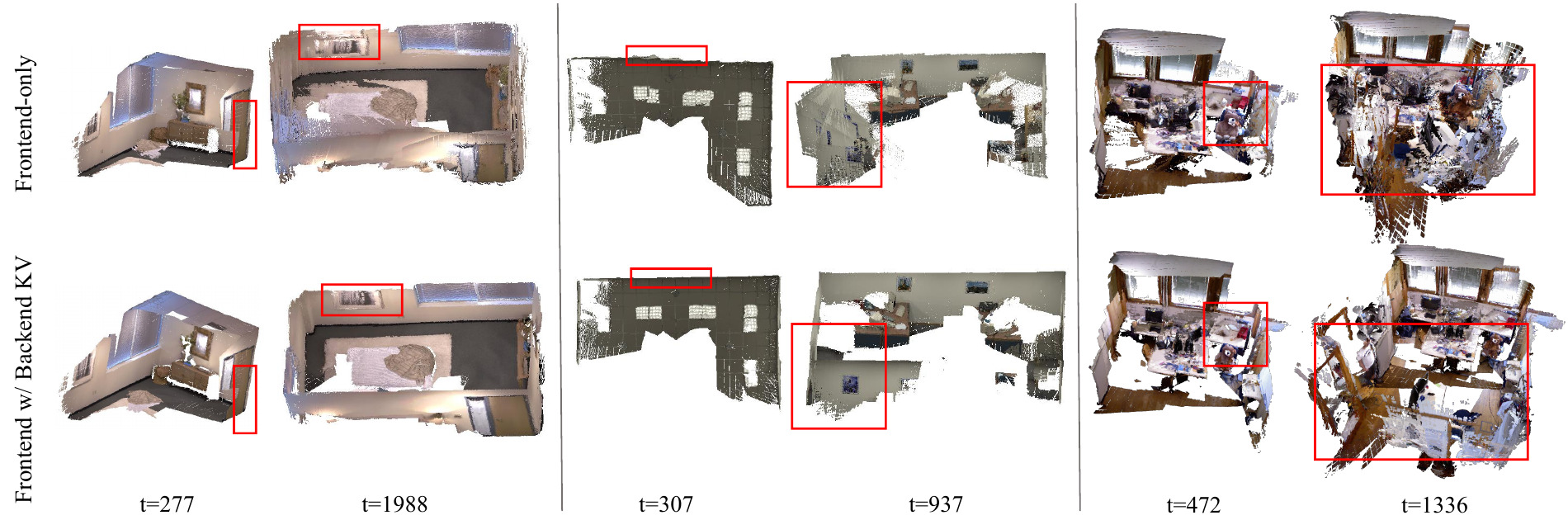} 
	\caption{Qualitative reconstruction comparison with and without backend assistance. The first row displays the frontend-only results at the corresponding timestamps, while the second row shows the results with the assistance of backend KV caches. }
	\label{fig:whole_seq}
\end{figure*}

We demonstrate intermediate results on some of the most challenging sequences in datasets, including Replica-room1, ICLNUIM~\cite{handa2014benchmark}-ofkt1, and TUM-room, all of which are inside-out captures of an indoor environment as illustrated in~\cref{fig:whole_seq}.
Initially, the errors in the frontend-only results are relatively small, as highlighted by the red windows. However, as time progresses, the frontend-only reconstruction becomes significantly distorted.
This distortion arises because the frontend-only \textbf{accumulates errors over time}, leading to substantial inaccuracy in the later stages.
In contrast, our model, which incorporates a backend, \textbf{maintains consistency throughout the entire process}, effectively mitigating these issues. 

\subsubsection*{How does the backend benefit from the frontend?}

\begin{figure*}[t!]
    \centering
    \begin{subfigure}[t]{0.3\textwidth}
        \centering
        \includegraphics[height=1in]{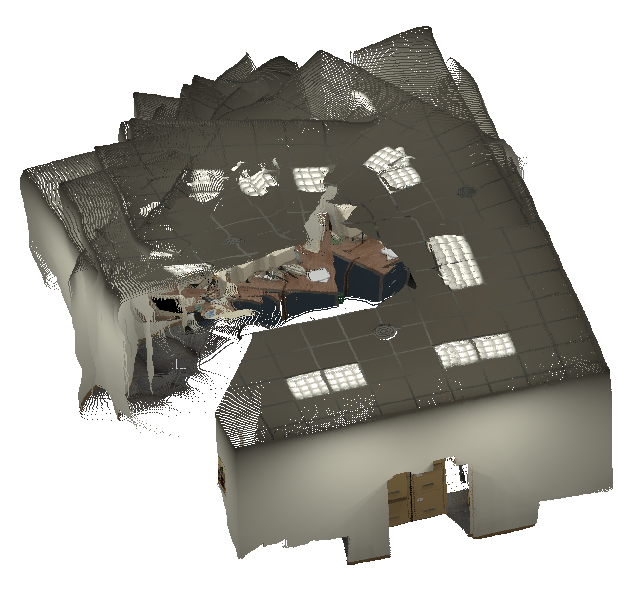}
        \caption{VGGT}
    \end{subfigure}%
    ~
    \begin{subfigure}[t]{0.3\textwidth}
        \centering
        \includegraphics[height=1in]{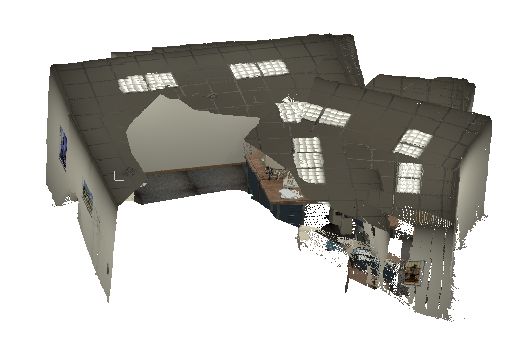}
        \caption{Pi3}
    \end{subfigure}%
    ~ 
    \begin{subfigure}[t]{0.3\textwidth}
        \centering
        \includegraphics[height=1in]{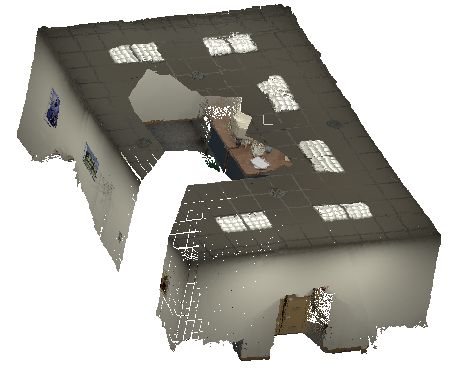}
        \caption{SLAM-Former}
    \end{subfigure}
    \caption{Qualitative reconstruction comparison on ICL-NUIM ofkt1. The results from VGGT, Pi3, and ours are displayed from left to right, respectively. 
	Both VGGT and Pi3 suffer from pose drift, leading to geometric inaccuracies, while our method demonstrates consistent and accurate reconstruction.}
    \label{fig:ablation2}
\end{figure*}

To answer this question, we conduct a demonstration using the ICL-NUIM scene, ofkt0 sequence.
As shown in~\cref{fig:ablation2}, the left two images display the results of VGGT and Pi3 when all keyframe images are used as input, without any sequential information.
The right image shows our result.
It is evident that without the sequential information provided by our frontend, 
VGGT and Pi3 produce disordered reconstructions.
In contrast, our backend leverages the implicit order provided by the frontend to achieve a more coherent and accurate reconstruction.
\subsection{Speeding up with KV Pruning}
\label{sec:exp:tokenprune}

\begin{table}[b]
\centering
\caption{Reconstruction evaluation on 7-Scenes~\cite{glocker2013real-time} (unit: m). Retained ratios indicate the fraction of the original KV kept after pruning. }
\label{tab:7scenes_recon_prune}
        \vspace{-3mm}

\setlength{\tabcolsep}{2.0pt}
\scriptsize
\renewcommand{\arraystretch}{\spacebetweentablerow}
\begin{tabular}{lcccc}
\toprule
Retained Ratio & \multicolumn{4}{c}{7-Scenes} \\ \cmidrule(lr){2-5}
& ATE\,$\downarrow$ & Acc.\,$\downarrow$ & Complet.\,$\downarrow$ & Chamfer\,$\downarrow$ \\
\midrule

100\% & 0.042 & 0.017 & 0.036 & 0.026 \\
50\% & 0.041 & 0.017 & 0.036 & 0.027 \\
25\% & 0.042 & 0.018 & 0.037 & 0.027 \\
12.5\% & 0.042 & 0.020 & 0.038 & 0.029 \\
6.25\% & 0.055 & 0.026 & 0.042 & 0.034 \\
\midrule
log KV pool & 0.042& 0.017& 0.037& 0.027\\
sqrt KV pool& 0.041& 0.017& 0.037& 0.027 \\
\bottomrule
\end{tabular}
\vspace{0.7em}

\end{table}

\begin{figure}[t!]
\centering
\begin{subfigure}[b]{0.49\textwidth}
    \centering
    \includegraphics[width=\textwidth]{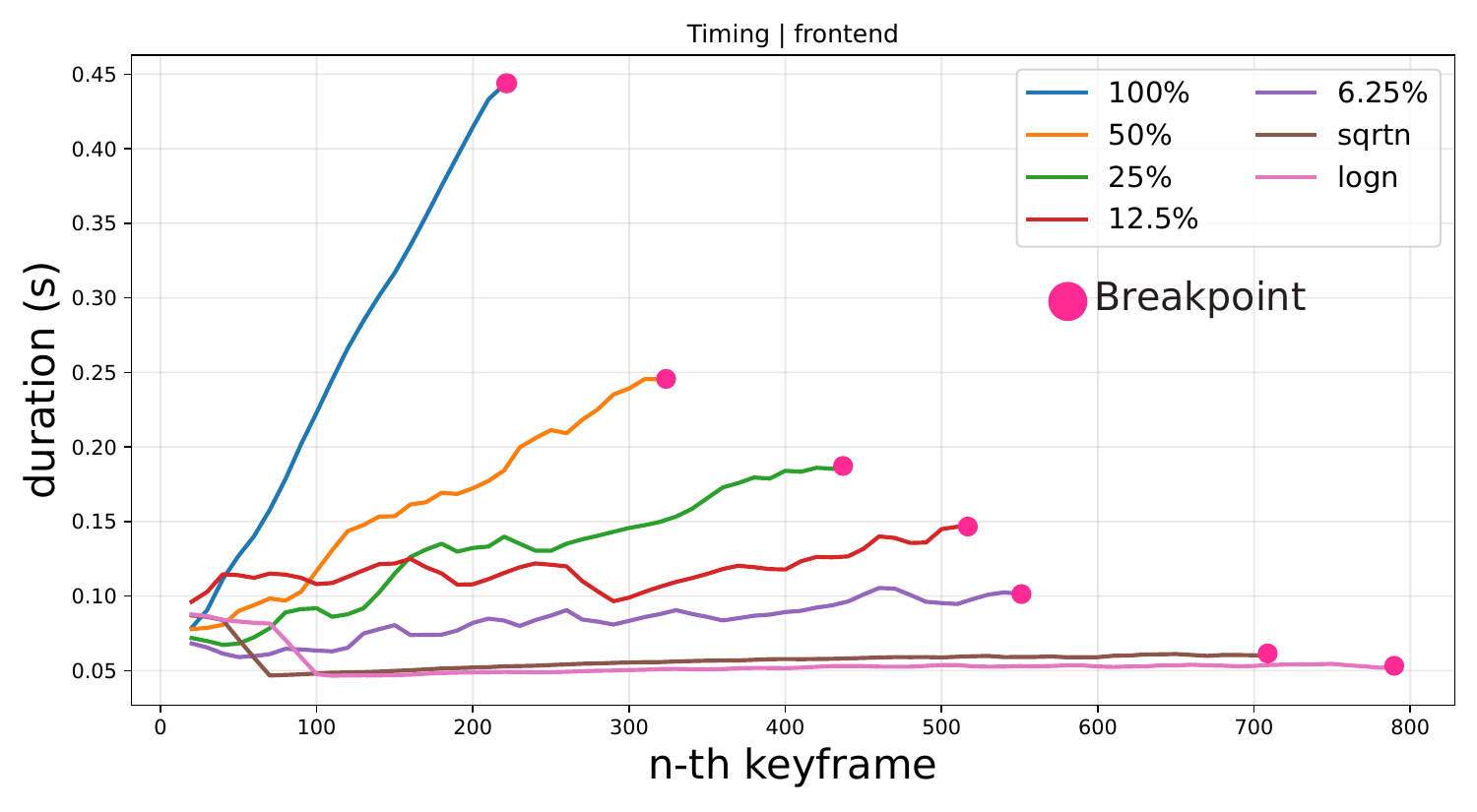}
    \caption{Frontend runtime}
\end{subfigure}
\hfill
\begin{subfigure}[b]{0.49\textwidth}
    \includegraphics[width=\textwidth]{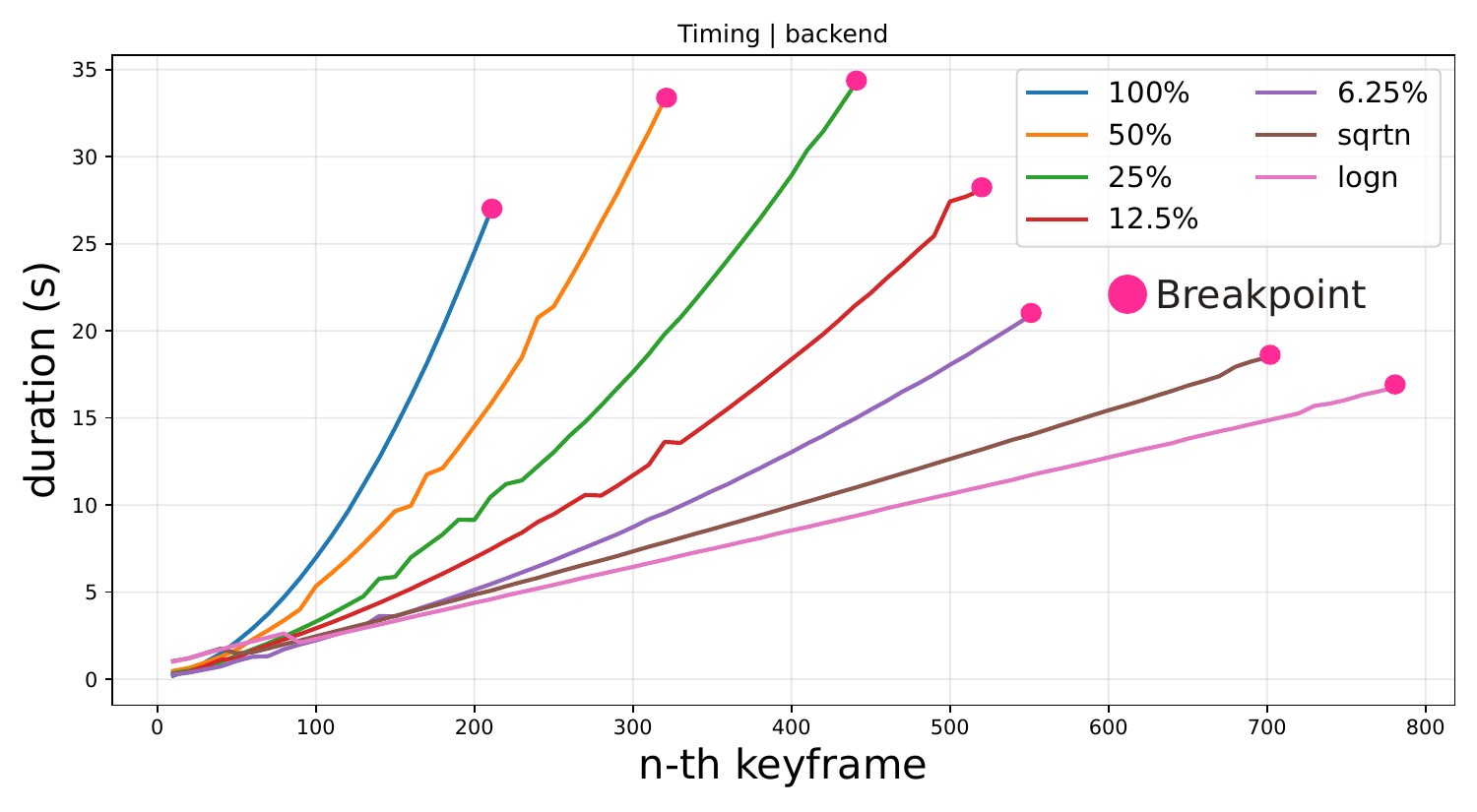}
    \caption{Backend runtime}
\end{subfigure}
\caption{Frontend and backend runtime on n-th keyframes.}
\label{fig:Frontend_and_backend_runtime}
\end{figure}

We evaluate the KV pruning 
on both runtime efficiency and reconstruction quality. As shown in Figure~\ref{fig:Frontend_and_backend_runtime} and Table~\ref{tab:7scenes_recon_prune}, the method delivers substantial gains in frontend and backend runtime while incurring only negligible degradation in reconstruction accuracy down to a 12.5\% retention ratio.
Noticeable degradation only appears when the retained information becomes overly sparse. Only then does the accuracy-efficiency trade-off begain to emerge.

This redundancy also suggests that information growth is sub-linear rather than linear. 
So our model has the potential to further reduce time complexity to for scalability.
That KV-retention schedule can change hyperparameter from const-ratio to dynamic-ratio, maintaining the performance (dynamic log- or sqrt-size KV pool in~\cref{tab:7scenes_recon_prune}) with $\mathcal{O}(nlog(n))$ or $\mathcal{O}(n^{1.5})$ time complexity. 
The time curves in~\cref{fig:Frontend_and_backend_runtime} demonstrate the effectiveness of KV Pruning. For each curve, the final point indicates the breakpoint before reaching the memory overhead.

\section{Conclusion}
In this work, we introduce SLAM-Former, putting full SLAM capability into a single transformer.
With alternating incremental frontend processing and global backend processing, SLAM-Former enables the frontend and backend to cooperate and enhance each other, resulting in an overall improvement.
Experiments demonstrate that SLAM-Former significantly outperforms traditional geometry foundation-based SLAM methods in both tracking and reconstruction. 
Furthermore, it achieves highly competitive tracking performance and vastly superior reconstruction compared to traditional methods on real-world data.

\section*{Acknowledgements}
This work is supported by the Beijing Municipal Science and Technology Program (Z251100008125011).

%
%
\bibliographystyle{splncs04}
\bibliography{main}
\end{document}


\maketitle

\setcounter{section}{0}
\renewcommand{\thesection}{\Alph{section}}
\renewcommand{\thetable}{\Alph{section}\arabic{table}}
\renewcommand{\thefigure}{\Alph{section}\arabic{figure}}
\renewcommand{\theequation}{\Alph{section}\arabic{equation}}

\section{Execution Latency}
\label{sec:execution_speed}
\setcounter{table}{0}
\setcounter{figure}{0}
We evaluate the computational time of two versions of our model: (1) a $518px$ version (default) and (2) a $224px$ version for lower computational cost.

Since our model is trained on images whose long side is $518$ (or $224$) pixels, all input images are resized accordingly.

We evaluate runtime with baselines on 7scenes dataset. The reported FPS is an average across all scenes. 
\begin{table}[htbp]
	\centering
        \caption{Average runtime cost on 7 scenes dataset.
		  }\label{tab:time}
		\resizebox{\textwidth}{!}{
	\begin{tabular}{l|@{\hspace{.2em}}c@{\hspace{.5em}}c@{\hspace{.5em}}c@{\hspace{.5em}}c} 
		\hline
		 Method &VGGT-SLAM & MASt3R-SLAM & \textbf{SLAM-Former} & \textbf{SLAM-Former} (224px)\\
      FPS &49.4&10.4&15.7 &25.8
           \\
		\hline
	\end{tabular}
    }
\end{table}

The $518px$ version is capable of running in real time.
The $224px$ version achieves faster execution with more than $25$ FPS. 

Note that the above evaluation does \textbf{not} include \textbf{KV pruning}, which could further accelerate both the frontend and backend by $2\times$, $4\times$, $8\times$, or more.



\section{Resolving Hard Scenes by Data Scaling}
\label{sec:data_scaling}
\setcounter{table}{0}
\setcounter{figure}{0}

Our proposed end-to-end SLAM model can handle complex and challenging scenes by incorporating appropriate training data.
Unlike traditional SLAM systems that rely on hand-crafted pipelines, our model learns geometry, tracking, and loop-closure cues directly from data.
In this framework, training replaces intricate engineering: by simply providing paired sensor sequences, the model learns to produce an accurate and real-time SLAM system without manual tuning or complex pipeline design.

One important scenario and example is in robotic navigating scenes, where Dense SLAM models often struggle due to large-scale and complex environments.
After learning from relevant training data, our method achieves robust performance in such environments.
To facilitate this, we leverage a robotic simulation environment to generate synthetic multi-room training data.

\subsection{Data Collection}
Specifically, we generate training sequences using Habitat-Sim~\cite{puig2023habitat3}, a simulator designed for large-scale indoor navigation research. It enables efficient rendering of high-fidelity RGB-D observations while providing accurate ground-truth information, including intrinsic and extrinsic parameters.



For scene assets, we employ the Habitat-Matterport 3D research dataset (HM3D)~\cite{ramakrishnan2021hm3d}, which contains 216 semantically annotated 3D indoor environments with diverse layouts. Navigation trajectories in these environments naturally span multiple rooms and cover long spatial distances, providing training sequences with rich spatial and temporal coverage.

\begin{figure*}[t]
    \centering
    \includegraphics[width=1\linewidth]{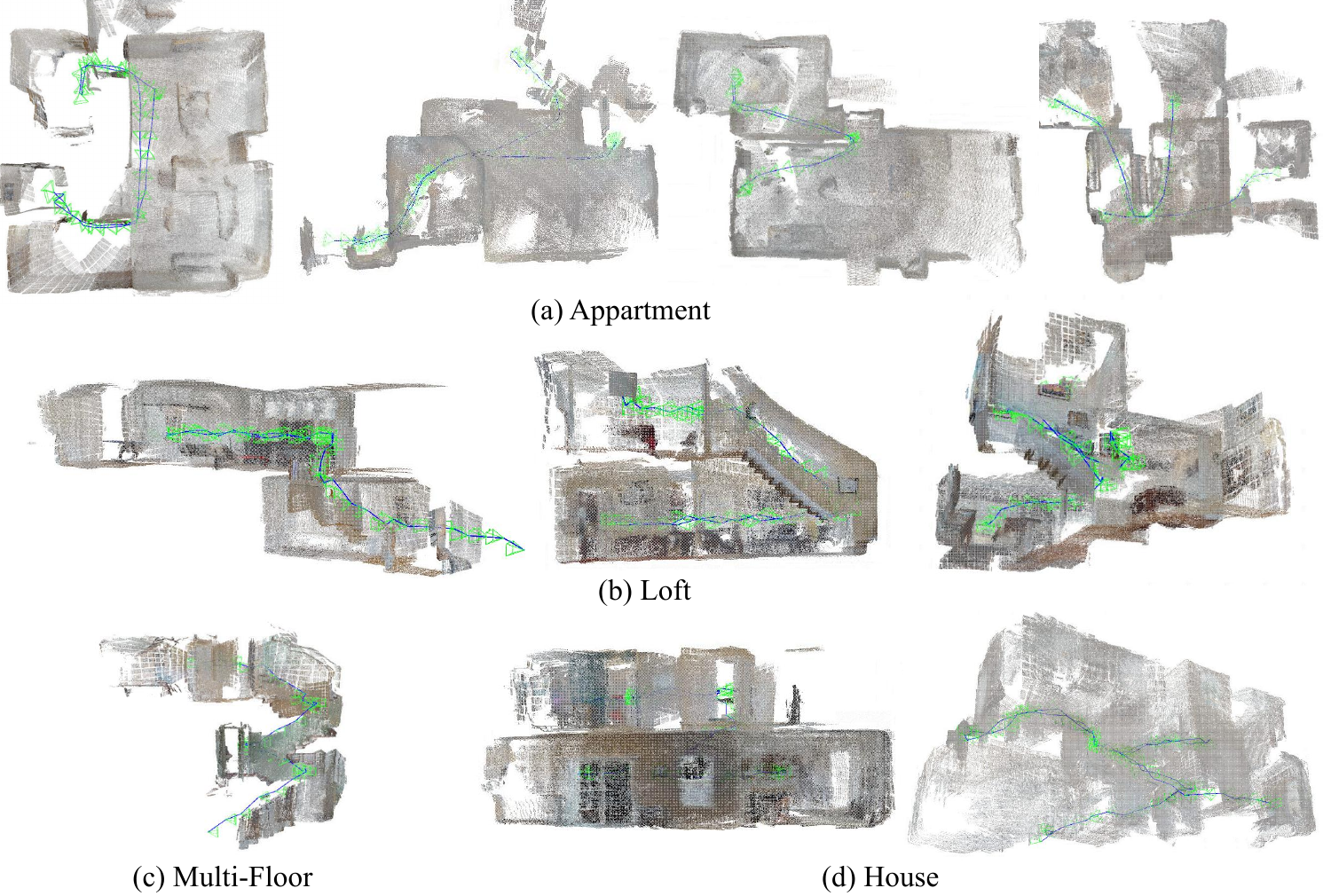}
    \caption{Qualitative results of SLAM-Former on a large-scale scene. The visualization shows predicted pointmaps across all frames, with the blue path representing the estimated camera trajectory and green frustums denoting camera views.}
    \label{fig:hm3d_val}
\end{figure*}

During data generation, each scene mesh is loaded into the simulator, and feasible start and goal locations are sampled according to semantic annotations, with the constraint that the shortest-path distance is at least 10 meters. A navigation trajectory is then generated using a shortest-path planner, producing a sequence of poses that traverses multiple regions of the environment.

To better simulate real-world camera motion, we further introduce random rotational and translational perturbations to keyframe poses before re-rendering observations. In total, we generate approximately 300 long free-camera trajectories from scenes in the HM3D training split. Each trajectory contains hundreds of keyframes and typically spans over 10 meters between the start and end positions. This large collection of long-horizon navigation sequences provides challenging multi-room training data for learning robust SLAM representations.

\subsection{Training and Evaluation on Large-scale Scenes}

For training, we adopt an input resolution of $224$ for SLAM-Former for computational efficiency.
This configuration reduces the memory footprint and accelerates training, allowing each batch to include sequences of $32$ keyframes.

We evaluate the model on a subset of scenes from the HM3D validation and test splits.
The scenes are categorized into four types with increasing difficulty: Apartment, Loft, Multi-Floor, and House.
The latter two explicitly capture multi-floor and multi-room complexities, which present significant challenges to SLAM systems, such as accumulated drift.
These challenging scenarios highlight the key advantage of our approach: the proposed end-to-end framework learns robust SLAM system directly from data under such conditions, without relying on hand-crafted pipelines.
Qualitative evaluations are conducted on these scenarios, with results illustrated in~\cref{fig:hm3d_val}.

The real-time mapping and localization results exhibit strong geometric consistency across all difficulty levels. 
Notably, even in the challenging Multi-Floor and House scenarios, no visible inconsistency is observed.

We also added evaluations on large scale HM3D and EuRoC MAV in~\cref{tab:hm3d_recon} and~\cref{tab:euroc_recon},
where our method performs overall better than MASt3R-SLAM on both tracking and mapping.
\begin{table}[htbp]
	\centering
    \vspace{-.3cm}
    \caption{Evaluation on HM3D.}
	\label{tab:hm3d_recon}
	\setlength{\tabcolsep}{2.0pt}
	\scriptsize
	\begin{tabular}{lccccc}
		\hline
		Method &ATE\,$\downarrow$ & Acc.\,$\downarrow$ & Complet.\,$\downarrow$ & Chamfer\,$\downarrow$ \\
		\hline
		MASt3R-SLAM & 0.655 &0.302&1.605&0.954\\
        \textbf{Ours} &\textbf{0.188}&\textbf{0.130}&\textbf{0.110}&\textbf{0.120}\\
		\hline
	\end{tabular}
        \vspace{-.3cm}
\end{table}
\begin{table}[htbp]
	\centering
        \vspace{-.3cm}
    	\caption{Evaluation on EuRoC.}
	\label{tab:euroc_recon}
	\setlength{\tabcolsep}{2.0pt}
	\scriptsize
	\begin{tabular}{lcccc}
		\hline
		Method & ATE\,$\downarrow$ & Acc.\,$\downarrow$ & Complet.\,$\downarrow$ & Chamfer\,$\downarrow$ \\
		\hline
		MASt3R-SLAM & 0.164 & 0.108 & 0.072&0.090\\
             \textbf{Ours} & \textbf{ 0.150}& \textbf{0.063}&\textbf{0.071} & \textbf{0.067}\\
		\hline
	\end{tabular}
\vspace{-.3cm}
\end{table}

\section{Additional Results}
\label{sec:additional_results}
\setcounter{table}{0}
\setcounter{figure}{0}

\subsection{KV-pruning Test on all Datasets}
We evaluate log-pruning and sqrt-pruning on all datasets as in~\cref{tab:kvprune}, where both demonstrate competitive results.

\begin{table}[htbp]
\centering
\caption{Evaluation on Replica, TUM and 7Scenes.}
\label{tab:kvprune}
\resizebox{\columnwidth}{!}{
\begin{tabular}{l|ccc|c|cccc}
\hline
&\multicolumn{3}{c|}{Replica}&\multicolumn{1}{c|}{TUM}&\multicolumn{4}{c|}{7Scenes}\\
\hline
\textbf{Method} &ATE&Acc.&Comp.&ATE&ATE&Acc.&Comp.&Chamf.\\
\hline
Ours&\textbf{0.030}&\textbf{2.09}&\textbf{1.56}&\textbf{0.039}&\textbf{0.042}&\textbf{0.017}&\textbf{0.037}&\textbf{0.027}\\
Ours (log KVpool)&0.034 &2.31&1.71&0.044&0.041&0.017&0.036&0.026\\
Ours (sqrt KVpool)&0.032&2.33&1.71&0.044&0.041&0.017&0.037&0.027\\
\hline
\end{tabular}
}
\vspace{-.5cm}
\end{table}

\subsection{Visual Comparison with MASt3R-SLAM}
\begin{figure}[htbp]
  \centering
  \includegraphics[width=\linewidth]{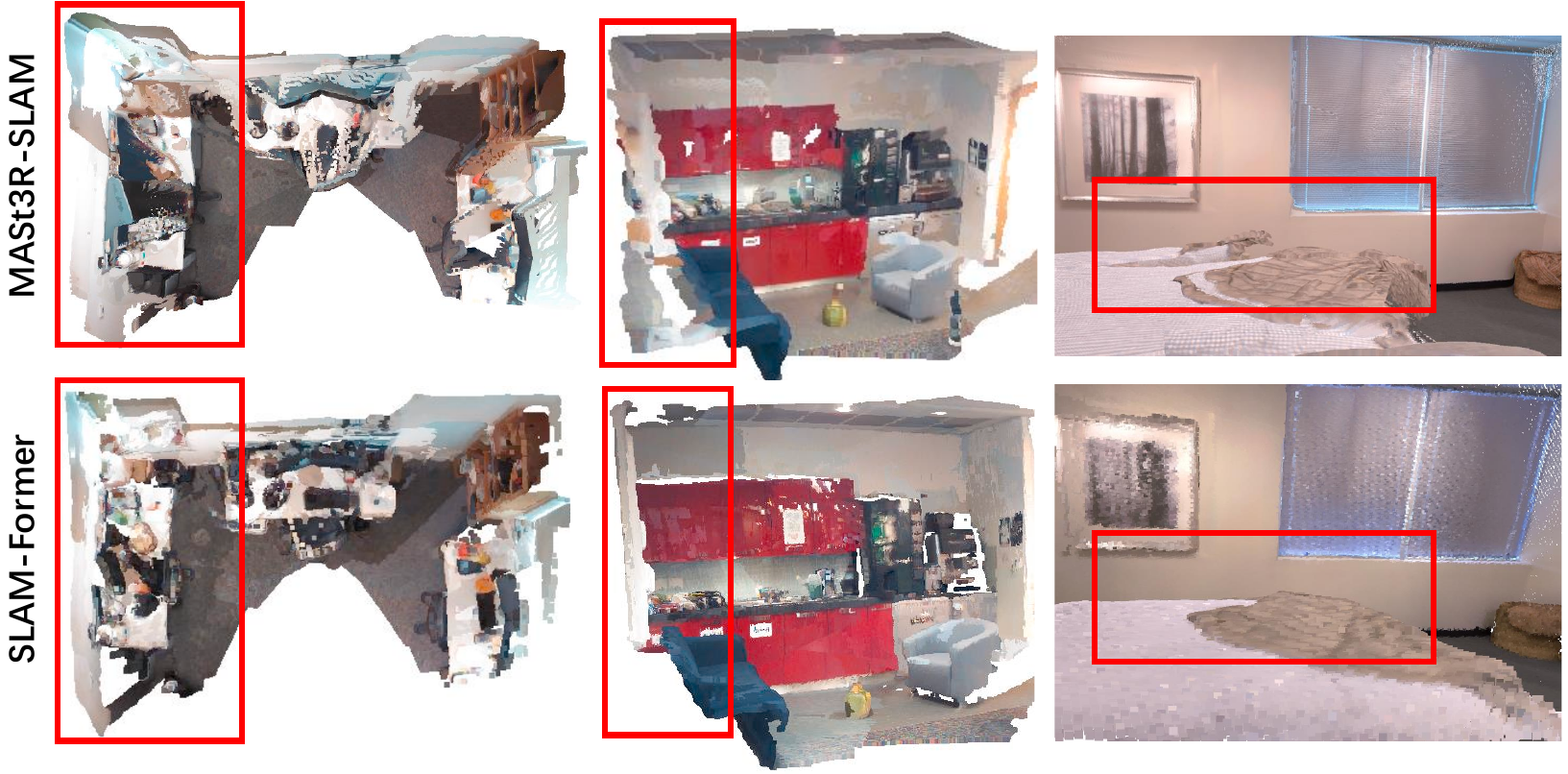}
  \caption{Qualitative comparison with MASt3R-SLAM.}
   \label{fig:mast3rslam-vis}
\end{figure}

We added qualitative comparison with MASt3R-SLAM in~\cref{fig:mast3rslam-vis}.
Our method achieves a more well-structured mapping.

\subsection{Testing on outdoor Scenes}

\begin{figure}[htbp]
    \centering
        \includegraphics[width=.4\linewidth]{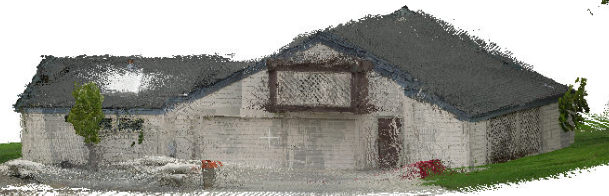}
        \includegraphics[width=.4\linewidth]{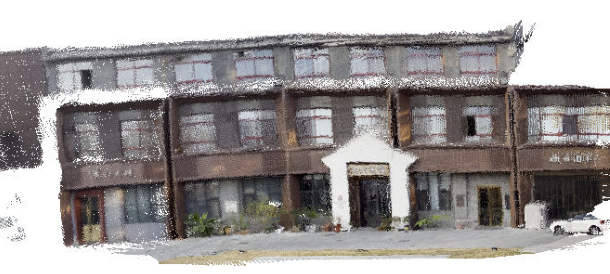}
    \caption{Qualitative results of SLAM-Former on on Tanks-and-Temple~\cite{Knapitsch2017} and DL3DV~\cite{ling2024dl3dv} scenes.  The visualization shows predicted pointmaps across all frame.}
    \label{fig:outdoor}
\end{figure}

Besides of our evaluation on the widely used indoor benchmarks, we also test on some outdoor scenes.
As demonstrated in~\cref{fig:outdoor}, our model works well on outdoor house and buildings.

\section{Ablation on SLAM Hyperparameters}
\setcounter{table}{0}
\setcounter{figure}{0}
We further conduct ablation study to investigate the impact of two SLAM hyperparameters: $T$, which controls the backend frequency (every $T$ keyframes), and $\tau$, the translation threshold for keyframe detection.

\begin{table}[htbp]
    \centering
    \caption{Ablation study on the effect of backend frequency $T$ on the 7-Scenes dataset. 
    ATE: Absolute Trajectory Error; Acc.: Accuracy; Comp.: Completeness; Chamfer.: Chamfer Distance.}
    \label{tab:ablation_T}
    \begin{tabular}{l | cccc}
    \toprule
        7scenes & ATE & Acc. & Comp. & Chamfer. \\ \hline
        $T=5$  & 0.044 & 0.017 & 0.037 & 0.027 \\
        $T=10$ & 0.042 & 0.017 & 0.037 & 0.027 \\
        $T=20$ & 0.043 & 0.017 & 0.037 & 0.027 \\
        $T=50$ & 0.046 & 0.018 & 0.038 & 0.028 \\
    \bottomrule
    \end{tabular}
\end{table}

As shown in~\cref{tab:ablation_T}, varying the backend period $T$ has only minor effects on final performance on the 7-Scenes dataset. Both tracking accuracy and mapping quality exhibit only small variations across different settings. 

\begin{table}[htbp]
    \centering
    \caption{Ablation study on the effect of the translation threshold $\tau$ for keyframe detection on the 7-Scenes dataset. 
    ATE: Absolute Trajectory Error; Acc.: Accuracy; Comp.: Completeness; Chamfer.: Chamfer Distance.}
    \label{tab:ablation_tau}
    \begin{tabular}{l | cccc}
    \toprule
        7scenes & ATE & Acc. & Comp. & Chamfer. \\ \hline
        $\tau=0.05$ & 0.044 & 0.015 & 0.034 & 0.025 \\
        $\tau=0.1$  & 0.042 & 0.017 & 0.037 & 0.027 \\
        $\tau=0.15$ & 0.045 & 0.019 & 0.040 & 0.030 \\
        $\tau=0.2$  & 0.043 & 0.022 & 0.042 & 0.032 \\
    \bottomrule
    \end{tabular}
\end{table}

For keyframe detection, as shown in~\cref{tab:ablation_tau}, we find that a smaller threshold $\tau$  tends to produce better reconstruction results. This is because a lower translation threshold results in denser keyframe selection, thereby preserving richer geometric details in the reconstructed map. 
Meanwhile, the impact on tracking accuracy remains marginal, with only minor variations observed across different $\tau$ values.

{
    \small
    \bibliographystyle{splncs04}
    \bibliography{main}
}